%% file: paper.tex
\documentclass{article}
\usepackage{byod_preprint,times}
\input{math_commands.tex}

\usepackage{booktabs}
\usepackage{graphicx}
\usepackage{hyperref}
\usepackage{microtype}
\usepackage{url}
\usepackage{xurl}
\hypersetup{pdftitle={BYOD: Build Your Own DLM - Efficient Autoregressive-to-Diffusion Language Model Conversion}, pdfauthor={Ruurd J. A. Kuiper; Robert A. Bagheri}, colorlinks=true, linkcolor=blue, citecolor=blue, urlcolor=blue}

\title{BYOD: Build Your Own DLM - Efficient Autoregressive-to-Diffusion Language Model Conversion}

\iclrfinalcopy
\input{authors.tex}

\newcommand{\method}{BYOD}
\newcommand{\na}{--}

\begin{document}
\maketitle

\begin{abstract}
Unlike autoregressive (AR) models, diffusion language models (DLMs) permit parallel token updates, flexible generation, and bidirectional attention, but leading systems rely on costly pretraining or full-model fine-tuning. We present Build Your Own DLM (BYOD), a generalizable LoRA-only recipe for parameter-efficient AR-to-DLM conversion. Without architecture-specific tuning, BYOD induces diffusion generation in Gemma 2 9B, Llama 3.1 8B, Qwen2.5 7B, and Ministral 8B; to our knowledge, this is the first reported AR-to-DLM conversion of a Mistral model. The adapters comprise 4.1--5.8\% of final parameters while every base-model parameter remains frozen; each 25,000-iteration conversion needs at most 102.4M tokens, processed in 3.1--4.0 hours on one Google Colab GPU. All four models generate from fully masked sequences, use bidirectional context in a controlled attention test, and retain substantial parent-model knowledge. BYOD-Gemma is the strongest BYOD conversion on our nine-task evaluation, reaching 50.6\% average accuracy compared with 51.9\% for LLaDA 8B Instruct in the same harness. Open-generation quality at low network function evaluation (NFE) budgets remains the principal limitation: BYOD perplexity is 8.3--11.1 versus 4.1 for LLaDA and 2.2--3.0 for the AR parents when generating 128 tokens in 64 steps. To enable further experimentation and inspection, we provide a standalone training and evaluation implementation and interactive demos for all four conversions. BYOD thus provides a practical route for extending diffusion generation to other existing model families while identifying the gap that efficient conversion must next close.

\noindent\textbf{Code:} \url{https://github.com/RuurdKuiper/BYOD}.\quad\textbf{BYOD demos:}
\href{https://huggingface.co/spaces/Ruurd/byod-gemma-2-9b}{Gemma} $\mid$
\href{https://huggingface.co/spaces/Ruurd/byod-llama-3.1-8b}{Llama} $\mid$
\href{https://huggingface.co/spaces/Ruurd/byod-qwen2.5-7b}{Qwen} $\mid$
\href{https://huggingface.co/spaces/Ruurd/byod-ministral-8b}{Ministral}
\end{abstract}

\section{Introduction}

Autoregressive (AR) language models generate left to right. This simple learning target scales effectively, but fixes the generation order, restricts attention to prior tokens, and produces only one new token per step. Diffusion language models (DLMs) instead iteratively reconstruct multiple positions in a corrupted sequence using bidirectional context \citep{austin2021structured,sahoo2024simple,shi2024simplified}. Their number of function evaluations (NFE) can be chosen independently of output length, and early low-probability predictions can be revised. Training or adapting models for diffusive generation, however, has remained computationally intensive and expensive.

\begin{figure*}[t!]
\centering
\includegraphics[width=\textwidth]{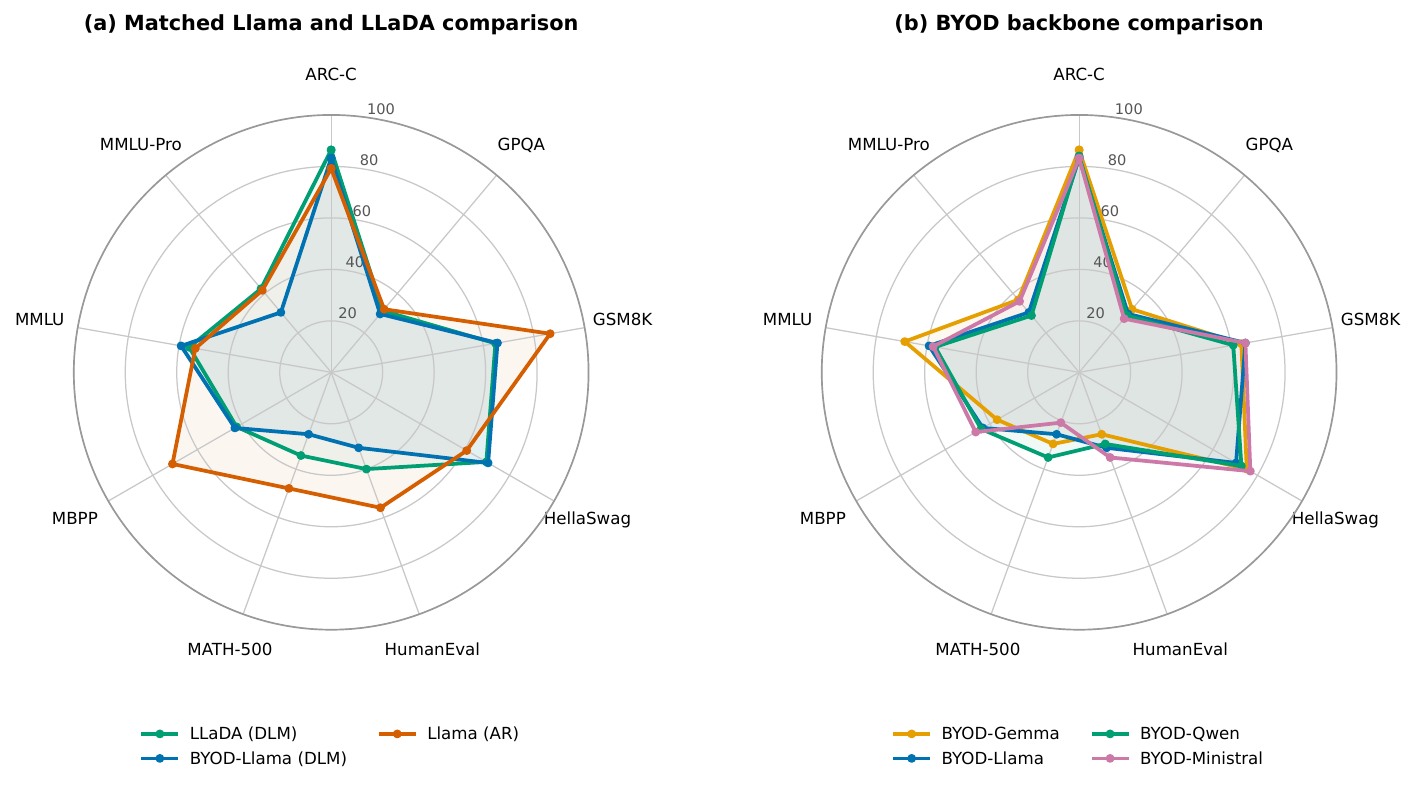}
\caption{Task-level accuracy profiles from the matched evaluation. (a) LLaDA, BYOD-Llama, and the unchanged Llama AR parent. (b) The four primary BYOD conversions.}
\label{fig:benchmark-radars}
\end{figure*}

Our primary aim is therefore a compute-efficient, reproducible recipe for converting AR models into DLMs without pretraining a diffusion foundation model for every architecture. This matters for three reasons. First, full diffusion pretraining is financially and environmentally costly: for example, under our standardized rental and electricity scenario, the reported LLaDA run corresponds to approximately \$675,000 and 44,100 kg CO$_2$e (Table~\ref{tab:compute}). It should be noted that these illustrative estimates are neither the authors' bill nor measured emissions; operational carbon depends strongly on utilization, datacenter efficiency, location, and energy supply \citep{strubell2019energy,patterson2021carbon}. Second, AR-to-DLM conversion in hours on one widely available cloud instance makes hands-on DLM research available beyond laboratories with multi-node clusters. Third, applying the same inexpensive intervention to multiple existing model families makes it feasible to study how architecture and parent-model choice affect diffusion generation without repeating foundation-model pretraining for every comparison.

We study this question with \method{}, a generalizable conversion and evaluation pipeline for four instruction-tuned decoder families: Gemma~2~9B, Llama~3.1~8B, Qwen2.5~7B, and Ministral~8B. We selected the latest, comparably sized text-only instruction checkpoints to avoid modality-specific confounding. Every backbone receives the same bidirectional attention intervention, LoRA target modules, and masked-diffusion objective. This paired design separates conversion quality from the strength of the starting checkpoint. We directly compare our models against LLaDA~8B Instruct because its standard masked-diffusion objective, instruction-tuning setup, and overlapping benchmark suite make it the cleanest external reference baseline. We additionally report published DiffuLLaMA, Dream-Instruct, iLLaDA-Instruct, and Efficient-DLM values, but keep them protocol-separated: their additional attention, masking, scheduling, data, and inference techniques would confound a controlled comparison. 

Our contributions are:
\begin{itemize}
    \item a compute-efficient recipe that converts four instruction-tuned AR model families to masked diffusion using only a few single-GPU hours and LoRA parameter updates;
    \item a paired evaluation of each diffusion model and its frozen AR parent across general knowledge, reasoning, mathematics, and code benchmarks;
    \item a controlled analysis of training duration, denoising-step count, validation loss, perplexity and distinct n-grams; and
    \item a standalone training and evaluation implementation and interactive demos for all four models, supporting accessible DLM conversion and qualitative evaluation.
\end{itemize}

\section{Related work}

Discrete diffusion models corrupt and reconstruct categorical sequences rather than adding Gaussian noise. D3PM formalized categorical corruption with transition matrices, including an absorbing process in which tokens transition to a dedicated mask state \citep{austin2021structured}. MDLM and contemporaneous work showed that this masked formulation admits a simple noise-weighted cross-entropy objective, while RADD reparameterized the predictor in terms of conditional distributions over clean tokens \citep{sahoo2024simple,shi2024simplified,ou2024absorbing}. These approaches form the basis of masked diffusion language models (MDLMs). Uniform (or multinomial) diffusion, which instead corrupts tokens with samples from the ordinary vocabulary, is another option \citep{hoogeboom2021multinomial}; we use the simpler absorbing-mask denoising objective employed by LLaDA.

\begin{table*}[t]
\caption{Incremental training scale and standardized resource estimates for comparable diffusion systems. GPU-hours marked $^\dagger$ are estimated from published model/data scale; unmarked values are reported. We stress that cost and CO$_2$e are estimates purely for illustrative purposes, based on a standardized operational scenario (Appendix~\ref{app:cost}), not reported values. NR: insufficient information.}
\label{tab:compute}
\centering
\scriptsize
\resizebox{\textwidth}{!}{%
\begin{tabular}{llllrrr}
\toprule
System & Diffusion-stage data & Updated weights & Hardware & GPU-h & Cost (USD) & CO$_2$e \\
\midrule
\method{} (ours) & $\leq$102.4M tokens & LoRA & 1$\times$RTX Pro 6000 & 3.2 & \$11 & 0.9 kg \\
DiffuLLaMA 7B & 65B tokens & Full & 64$\times$GH200 & 1,917$^\dagger$ & \$9,951 & 0.65 t \\
LLaDA 8B & 2.3T tokens & Full & H800 & 130,000 & \$674,830 & 44.1 t \\
Dream 7B & 580B tokens + SFT & Full & NR & 17,105$^\dagger$ & \$88,791 & 5.81 t \\
iLLaDA 8B & 12T tokens + SFT & Full & NR & 404,449$^\dagger$ & \$2,099,494 & 137.3 t \\
Efficient-DLM 8B & 500B tokens & Full & 128$\times$H100 & 16,852$^\dagger$ & \$87,479 & 5.72 t \\
Uno 8B & 7B tokens & LoRA & 64$\times$H200 & 3,840 & \$19,933 & 1.30 t \\
Fast-dLLM v2 7B & $\sim$1B tokens & Full & 64$\times$A100 & 768 & \$3,987 & 149 kg \\
\bottomrule
\end{tabular}}
\end{table*}

The authors of LLaDA were the first to scale masked diffusion to an 8B general-purpose language model \citep{nie2025llada}. Its architecture follows Llama~3.1~8B, but it was pretrained from scratch using 2.3T tokens and 130,000 GPU-hours. AR-to-diffusion conversion instead reuses knowledge from an AR checkpoint. DiffuLLaMA established this route through 65B tokens of full-model continued training at 7B scale \citep{gong2025diffullama}. Dream uses DiffuLLaMA's shifted-prediction initialization, continues Qwen2.5 for 580B tokens, and adds context-adaptive token-level noise rescheduling \citep{ye2025dream}. Efficient-DLM combines 500B tokens of continued training with blockwise attention and position-dependent masking \citep{fu2025efficientdlm}, whereas iLLaDA uses from-scratch diffusion pretraining at substantially larger scale \citep{nie2026illada}. Conversion has also reached larger sparse models through RND1 (30B parameters), LLaDA2.0 (16B and 100B), and DiffusionGemma (25.2B, using uniform diffusion) \citep{chandrasegaran2025rnd1,bie2025llada20,diffusiongemma2026}. Thus, Gemma-, Llama-, and Qwen-based diffusion models were already reported, but we found no prior diffusion conversion of Ministral~8B Instruct~2410.

Hybrid and blockwise models offer a different compromise between bidirectionality and deployability. Block Diffusion is autoregressive across blocks but diffusive within each block, enabling variable-length generation and inter-block KV caching \citep{arriola2025block}. Fast-dLLM v2 converts Qwen2.5 models using full-parameter fine-tuning with approximately 1B tokens and combines blockwise complementary masking with hierarchical caching \citep{wu2025fastdllmv2}. These systems retain more left-to-right structure than our fully bidirectional answer block, making them especially relevant to speed, caching, and arbitrary-length generation but not suited as direct controls for our intervention.

LAD first showed that LoRA can induce diffusion generation, using a hybrid objective that combines masking with token swaps, duplications, and span shifts \citep{kuiper2025lad,hu2022lora}. However, \method{} instead tests whether a simpler IID absorbing-mask objective can transfer reliably across model families. Other LoRA-based diffusion work addresses different goals: Uno learns diffusion-draft adapters while retaining the AR model as verifier, the dLLM framework uses LoRA to fine-tune already-diffusive checkpoints rather than for its Tiny-A2D conversion, and NaRA adapts existing diffusion models with noise-conditioned LoRA \citep{sahoo2026uno,zhou2026dllm,wang2026nara}. Thus, their LoRA components augment already diffusively generating models, rather than converting AR model families into standalone diffusion models.

Table~\ref{tab:compute} compares training budgets for similarly sized diffusion models. For AR-to-diffusion conversion methods, these budgets exclude the substantial earlier cost of training the AR parent; for from-scratch methods they include diffusion pretraining. Dollar and operational-emissions columns are rental-equivalent estimates under one documented pricing and energy scenario, not the authors' actual bills or measured footprints; Appendix~\ref{app:cost} gives the complete calculation and its limitations.

Our study has explicitly chosen to hold the adaptation mechanism as constant as the architectures permit, repeat it across four families, and benchmark each converted model against the exact AR checkpoint from which it originated. This paired comparison is essential because it allows us to compare our results directly with the prior capability of the parent model.

\section{Method}

\subsection{Masked-diffusion training objective}

The training objective teaches the adapted model to reconstruct masked response tokens while conditioning on an uncorrupted instruction.

Let $p$ be the instruction, $x_0=(x_0^1,\ldots,x_0^L)$ the clean response region, $A$ the answer positions, and $P$ the EOS-padding positions. For each example, we sample $t\sim\mathcal{U}[10^{-3},1]$ and independently mask each position in $A\cup P$ with probability $t$, yielding the corrupted sequence $x_t$ and masked set $M_t$. For either region $R\in\{A,P\}$, we calculate the same-position reconstruction loss
\begin{equation}
\ell_R(\theta)=-\frac{1}{t|R|}
\sum_{i\in M_t\cap R}
\log p_{\phi,\theta}(x_0^i\mid p,x_t).
\label{eq:region-loss}
\end{equation}
We then combine the answer and padding losses:
\begin{equation}
\mathcal{L}(\theta)=\mathbb{E}_{t,x_0,x_t}
\left[0.9\ell_A(\theta)+0.1\ell_P(\theta)\right].
    \label{eq:objective}
\end{equation}
Here, $p_{\phi,\theta}$ is the conditional token distribution, $\phi$ denotes the frozen parent parameters, and $\theta$ the LoRA parameters. The first terminating EOS is included as a genuine answer token; all subsequent EOS tokens are treated as padding. The factor $1/t$ corrects for the sampled masking rate. Prompt and chat-template tokens remain visible to the model and are not part of the loss. Training masks are resampled online; validation masks are fixed by example index and seed.

\begin{table}[t]
\caption{Diffusion model names matched to parent models, their trainable parameter counts, and measured time for 25,000 updates. All trainable parameters are rank-1024 Q/V LoRA weights.}
\label{tab:models}
\centering
\small
\resizebox{\columnwidth}{!}{%
\begin{tabular}{llrrrr}
\toprule
Diffusion model & AR parent & Total parameters & Trainable & Trainable (\%) & Time (h) \\
\midrule
BYOD-Gemma & Gemma 2 9B IT & 10.732B & 572.5M & 5.33 & 4.0 \\
BYOD-Qwen & Qwen2.5 7B Instruct & 7.939B & 323.0M & 4.07 & 3.1 \\
BYOD-Ministral & Ministral 8B Instruct 2410 & 8.511B & 490.7M & 5.77 & 3.4 \\
BYOD-Llama & Llama 3.1 8B Instruct & 8.466B & 436.2M & 5.15 & 3.2 \\
\bottomrule
\end{tabular}}
\end{table}

\subsection{Autoregressive-to-diffusion conversion}

To convert an autoregressive model to a diffusion-style model, we replace the causal mask in every layer with a bidirectional attention mask, so every prompt, answer, and EOS-padding position can attend to every position. We attach LoRA adapters to every self-attention query and value projection, with rank $r=1024$, scaling $\alpha=1024$, and dropout zero. The rest of the model architecture and output vocabulary remain unchanged. Backbone weights, token embeddings, the language-model head, biases, and normalization layers are frozen. The pretrained output head is also retained unchanged: whereas autoregressive training compares each position’s logits with the next token through shifted labels, BYOD removes this label shift and predicts the clean token at the same position from the bidirectionally attended, corrupted sequence. Each model reuses an existing single-token string from its tokenizer (\texttt{MASK} for Gemma, Llama, and Qwen; \texttt{<?>} for Ministral), avoiding vocabulary expansion.

We fix the LoRA rank across model families to retain one architecture-independent recipe. Matching trainable parameter counts would require model-specific ranks and would still not control for differences in depth, width, and attention design. This results in variation in trainable parameters, which is therefore a potential confounder in cross-family comparisons. Every conversion was trained on one NVIDIA RTX PRO 6000 GPU in Google Colab, without distributed parallelism; Table~\ref{tab:models} reports parameter counts and conversion times.

For deployment, each adapter can be folded into its parent projection as $W_{\mathrm{merged}}=W+(\alpha/r)BA$, where $W$ is the frozen projection and $A$ and $B$ are the learned low-rank matrices. This removes the adapter branch while retaining the bidirectional attention intervention. Appendix~\ref{app:merge} evaluates whether merging changes model behavior.

\subsection{Training data}

We construct a one-million-example mixture containing 45\% general instruction following, 18\% multiple-choice reasoning, 18\% mathematics, and 19\% code. Within these groups, the source weights are: Clean-Instruct-3M/T\"ulu~3/Alpaca-GPT4/Alpaca at 40/35/20/5 \citep{cleaninstruct3m,lambert2024tulu3,peng2023instructiongpt4,taori2023alpaca}; MMLU/HellaSwag/ARC-Easy at 73/25/2 \citep{hendrycks2021mmlu,zellers2019hellaswag,clark2018arc}; Orca-Math/GSM8K at 95/5 \citep{mitra2024orcamath,cobbe2021gsm8k}; and OpenCoder/MBPP at 99.7/0.3 \citep{huang2024opencoder,austin2021mbpp}. When applicable, only the training splits are used. We deduplicate normalized prompts, exclude exact matches with held-out benchmark prompts, and retain only examples that fit the 256-token training length. Each model receives the same shuffled examples, retokenized with its native chat template and tokenizer. Prompt tokens remain clean; only answer and EOS-padding positions can be corrupted.

The four primary models see the same dataset prefix. BYOD-Llama-50k uses the same fixed shuffle for a longer, 800,000-example run with newly sampled corruptions. Exact run sizes and optimization settings are given in Appendix~\ref{app:training-settings}.

\subsection{Inference}
Inference begins from a fixed-length, fully masked answer block. At each denoising step, the model predicts all masked positions, retains the most confident predictions, and remasks lower-confidence positions; accepted tokens remain unmasked. This can resolve several output tokens per network evaluation. Task-specific lengths, denoising budgets, temperatures, and EOS/EOT handling are given in Appendix~\ref{app:inference-settings}.

\begin{figure}[t]
\centering
\fbox{\begin{minipage}{0.92\columnwidth}
\small
\textbf{User:} Replace the masked token in the partial assistant response with exactly the word written after ``COPY:''. Copy the spelling exactly. Do not use a synonym or a related word.\par\smallskip
\textbf{Initial masked model input:}\par
\texttt{Answer: [MASK]. COPY: apple.}\par\smallskip
\textbf{Expected model response:}\par
\texttt{Answer: apple. COPY: apple.}
\end{minipage}}
\caption{Example input and expected reconstruction for the bidirectional attention test. Only the masked position is predicted.}
\label{fig:bidirectional-example}
\end{figure}

\section{Evaluation}
\subsection{Paired models and baselines}
We refer to the four conversions as BYOD-Gemma, BYOD-Qwen, BYOD-Ministral, and BYOD-Llama. For Llama, we distinguish the primary and longer-trained checkpoints as BYOD-Llama-25k and BYOD-Llama-50k. We also compare each model with its unchanged AR parent. LLaDA~8B Instruct serves as an external diffusion baseline in the same harness; results reported by other papers are presented separately because their protocols differ. We specifically use BYOD-Llama for the longer run and checkpoint-wise analysis because DiffuLLaMA, LLaDA, and iLLaDA use Llama or closely related Llama-style architectures, making it the clearest basis for comparison with prominent general-purpose DLMs \citep{gong2025diffullama,nie2025llada,nie2026illada}. We also compare BYOD-Llama-25k before and after folding its LoRA weights into the base model, to see if merging deteriorates the results. The nine standard task benchmarks use greedy token selection and low-confidence remasking, while open-ended generation uses temperature 0.7 and delayed EOS/EOT acceptance. Complete settings are reported in Appendix~\ref{app:inference-settings}.

\subsection{Task benchmarks}
Our benchmark set contains the validation/test sets of MMLU, MMLU-Pro, HellaSwag, ARC-Challenge, GSM8K, MATH-500, GPQA-Main, HumanEval, and MBPP \citep{hendrycks2021mmlu,wang2024mmlupro,zellers2019hellaswag,clark2018arc,cobbe2021gsm8k,hendrycks2021math,rein2023gpqa,chen2021codex,austin2021mbpp}. We score multiple-choice tasks by option label, GSM8K by its final number, MATH by its normalized final boxed expression, and code tasks by whether the generated program passes the supplied tests in an isolated subprocess.

\subsection{Open-ended generation}
We assess output fluency with perplexity and Distinct-1/2/3 calculated over answers on 100 open-ended questions. Standard masked diffusion does not directly provide the left-to-right sequence likelihood used for AR perplexity, and its usual evidence lower bound is not a proper test-time likelihood \citep{turok2026duel}. We therefore follow prior DLM work and use frozen Microsoft Phi-4 14B as an external perplexity scorer \citep{gong2025diffullama}. We pool negative log-likelihoods across all non-padding tokens of all generated answers before exponentiating, so every token has equal weight. Because this score does not measure diversity, we also report sliding model-token Distinct-1/2/3 percentages after removing special tokens. Higher Distinct-$n$ indicates less local repetition \citep{li2016diversity}; in our experiments, it exposed cases where perplexity improved because output became more repetitive rather than better.

\subsection{Bidirectional attention test}
We test whether the converted diffusion models allow bidirectional information flow with a simple right-context copy task (Appendix~\ref{app:bidirectional-infilling}). Most of the response is pre-filled, and the model must fill an earlier masked position by copying a word shown later in the response. A causal model cannot change its prediction at an earlier position based on a later token, so sensitivity to this clue directly tests functional bidirectional attention. Figure~\ref{fig:bidirectional-example} shows one test item and its expected reconstruction.

\section{Results}

\subsection{Benchmark accuracy and base-model dependence}

\begin{table*}[t]
\caption{Exact-match/pass@1 accuracy in our matched harness. D denotes a converted diffusion model and AR its unchanged parent. The Llama D-25k and D-50k columns share the same AR parent. Among common-harness diffusion columns, bold marks the best result and underline the best BYOD result (ties retained); published LLaDA is protocol-separated and excluded from highlighting.}
\label{tab:accuracy}
\centering
\scriptsize
\resizebox{\textwidth}{!}{%
\begin{tabular}{lccccccccccc}
\toprule
& \multicolumn{2}{c}{Gemma} & \multicolumn{2}{c}{Qwen} & \multicolumn{2}{c}{Ministral} & \multicolumn{3}{c}{Llama} & \multicolumn{2}{c}{LLaDA} \\
Task & D & AR & D & AR & D & AR & D-25k & D-50k & AR & Ours & Published \\
\midrule
\multicolumn{12}{c}{\textit{General knowledge and reasoning (multiple choice)}} \\
\midrule
ARC-C      & \textbf{\underline{.864}} & .888 & .840 & .912 & .832 & .768 & .832 & .824 & .792 & \textbf{.864} & .885 \\
GPQA       & \textbf{\underline{.320}} & .280 & .288 & .344 & .272 & .368 & .296 & .272 & .320 & .312 & .333 \\
HellaSwag  & .752 & .696 & .728 & .752 & \textbf{\underline{.768}} & .680 & .704 & .760 & .608 & .696 & .746 \\
MMLU       & \textbf{\underline{.688}} & .648 & .568 & .656 & .576 & .544 & .592 & .616 & .536 & .560 & .655 \\
MMLU-Pro   & \underline{.368} & .448 & .288 & .424 & .360 & .328 & .304 & .328 & .416 & \textbf{.424} & .370 \\
\midrule
\multicolumn{12}{c}{\textit{Math (open ended)}} \\
\midrule
GSM8K      & .640 & .872 & .608 & .920 & \textbf{\underline{.656}} & .872 & \textbf{\underline{.656}} & .616 & .864 & .648 & .694 \\
MATH-500   & .296 & .600 & \textbf{\underline{.352}} & .784 & .208 & .576 & .256 & .296 & .480 & .344 & .319 \\
\midrule
\multicolumn{12}{c}{\textit{Code generation (open ended)}} \\
\midrule
HumanEval  & .256 & .740 & .296 & .890 & \underline{.352} & .940 & .312 & .280 & .560 & \textbf{.400} & .494 \\
MBPP       & .368 & .740 & .440 & .760 & \textbf{\underline{.464}} & .720 & .432 & .440 & .712 & .424 & .410 \\
\midrule
Average    & \underline{.506} & .657 & .490 & .716 & .499 & .644 & .487 & .492 & .588 & \textbf{.519} & .545 \\
\bottomrule
\end{tabular}}
\end{table*}

Table~\ref{tab:accuracy} shows that BYOD-Gemma has the highest BYOD average and comes within 1.3 points of LLaDA in the matched harness. The AR parents perform better on average, but the conversion gap varies by task and backbone. Extending BYOD-Llama from 25,000 to 50,000 updates raises its average by 0.5 points, with gains on five tasks and declines on four.

Figure~\ref{fig:benchmark-radars} shows the largest gaps with AR in mathematics and code. On paired HumanEval/MBPP subsets, parsing failed for 28.0\%/8.8\% of BYOD-Llama-50k outputs versus 0\%/0\% for Llama AR (Appendix~\ref{app:code-errors}). LLaDA scores 2.6 points below its published average in our harness, illustrating protocol sensitivity; other external comparisons appear in Appendix~\ref{app:published-results}. The merged BYOD-Llama remains broadly similar across factual and open-generation measures (Appendix~\ref{app:merge}).

\subsection{Open-ended generation}
\label{sec:perplexity}

\begin{figure*}[t]
\centering
\includegraphics[width=\textwidth]{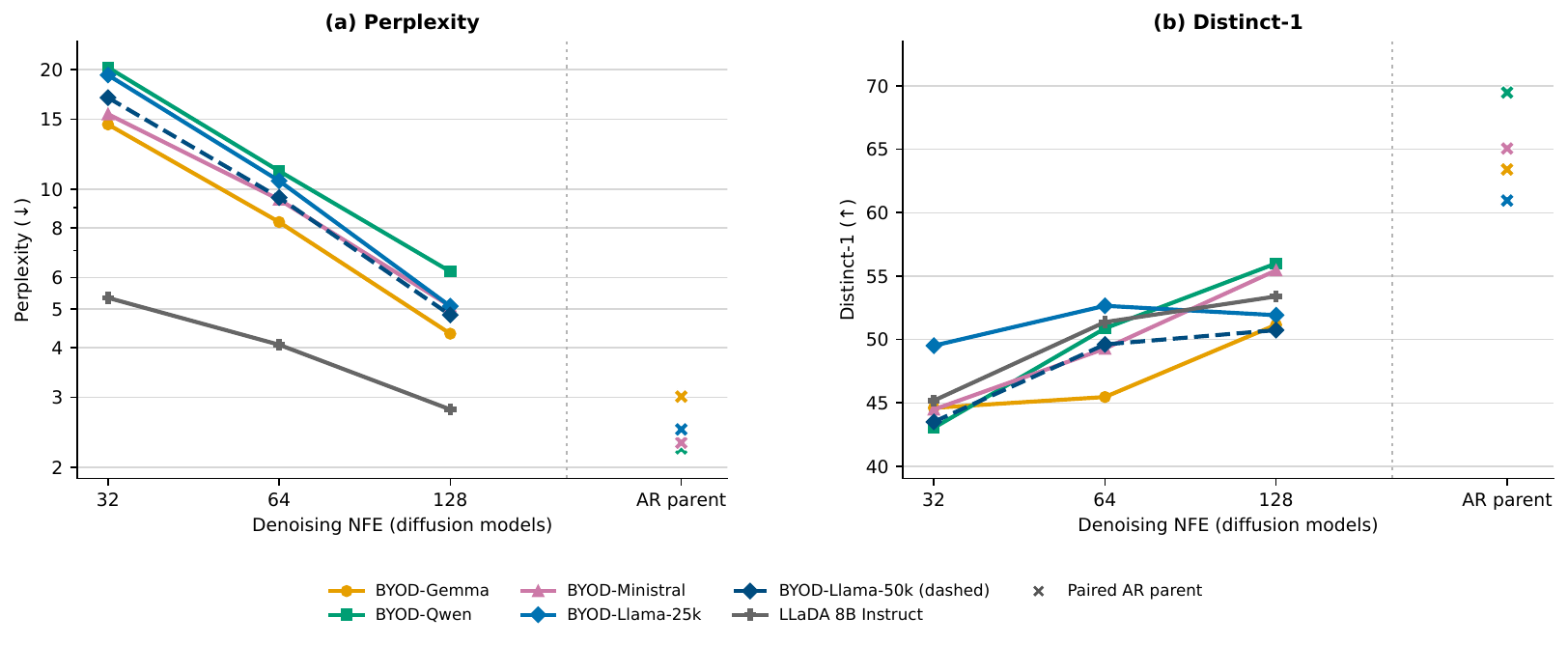}
\caption{Fluency metrics for different models with different NFE budgets. Lines connect each diffusion model's results at 32, 64, and 128 NFE; crosses to the right show the corresponding AR-parent reference. Perplexity is shown on a logarithmic axis.}
\label{fig:open-generation-budget}
\end{figure*}

Figure~\ref{fig:open-generation-budget} shows that perplexity and Distinct-1 generally improve with additional NFE. Appendix~\ref{app:open-generation-results} reports the exact values: across NFE budgets, converted models have 2.7--5.0 times their AR parents' perplexity and generally lower Distinct-$n$. For simplicity, we discuss Distinct-1 in the main text; appendix results for Distinct-1/2/3 show the same broad pattern. LLaDA also has lower perplexity than every BYOD model. Representative samples are provided in Appendix~\ref{app:demo-trajectories} to illustrate iterative refinement under different remasking strategies and NFE budgets.

\subsection{Further training}

\begin{figure*}[t]
\centering
\includegraphics[width=\textwidth]{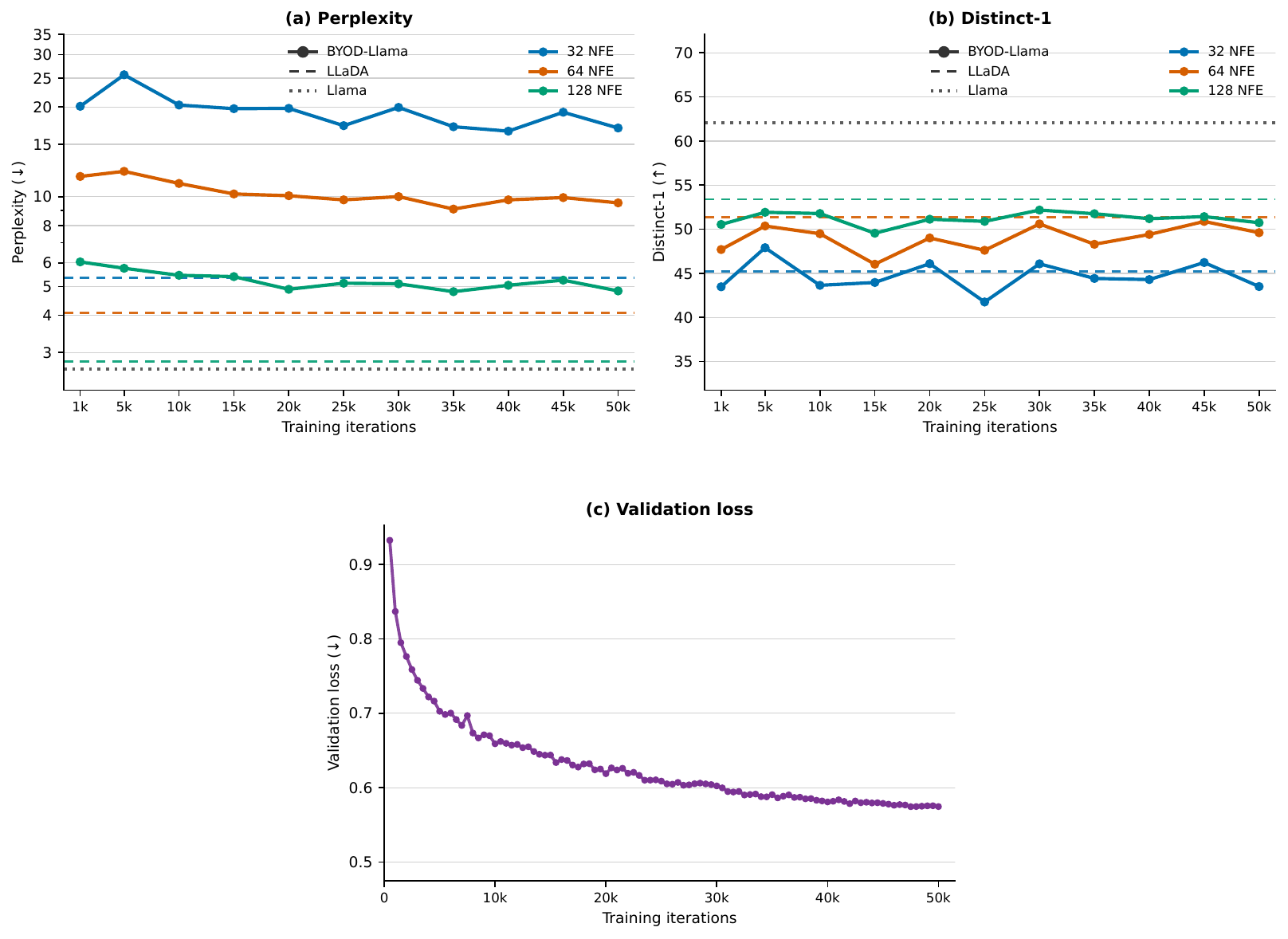}
\caption{BYOD-Llama-50k training diagnostics. (a) Token-weighted Phi-4 perplexity for 128-token generations, shown on a logarithmic y-axis. (b) Mean sliding model-token Distinct-1 for the same generations, reported as a percentage. Color-matched dashed lines are fixed LLaDA references; the dotted gray line is the unchanged Llama AR parent. (c) Validation loss.}
\label{fig:llama-training-diagnostics}
\end{figure*}

The benchmark results show small but uneven accuracy gain from further training. Figure~\ref{fig:llama-training-diagnostics} shows that validation loss decreases almost monotonically and begins to plateau toward the end of 50,000 training updates. Perplexity also decreases across all NFE budgets, but with larger fluctuations between checkpoints. Distinct-1 changes comparatively little and remains broadly stable. BYOD-Llama perplexity remains well above LLaDA, whose perplexity is in turn above the Llama AR baseline. In contrast, BYOD-Llama and LLaDA have similar Distinct-1 scores, while the AR baseline remains higher.

\subsection{Bidirectional attention test}
\label{sec:bidirectional-infilling}

Adding the later clue changed the greedy prediction on 100\% of test items for every converted model and LLaDA, confirming that all models use right-context information. BYOD-Gemma, BYOD-Ministral, and LLaDA copied the visible target exactly in 100\% of cases; BYOD-Llama-25k, BYOD-Llama-50k, and BYOD-Qwen reached 90\%, 70\%, and 20\%, respectively. Without the clue, exact-match accuracy was 0\% for every model. Appendix~\ref{app:bidirectional-infilling} reports the complete results and protocol.

\section{Discussion}

The central result is that high-rank attention-only LoRA can convert four AR models into DLMs in a few hours on a single GPU. The converted models do not surpass their AR parents on average, but do approach another DLM, LLaDA, on benchmark accuracy. This demonstrates that the conversion is feasible and effective, even if it does not fully close the performance gap. Several aspects of these findings warrant closer examination, and are discussed below.

Conversion quality varies by backbone: BYOD-Gemma performs best on benchmark tasks despite Qwen having the strongest AR parent, so parent capability alone does not determine conversion quality. Gemma's larger model and adapter budget may contribute, but cannot be separated here. The largest benchmark deficits relative to the AR parents occur in mathematics and code. In code, parser checks and qualitative inspection reveal syntactic errors, which do not occur in AR answers (Appendix~\ref{app:code-errors}). This is consistent with prior work showing that parallel token prediction in DLMs can produce syntactic inconsistencies \citep{hersche2026codila}.

Although generation under full NFE budgets generally produces coherent sentences, perplexity and Distinct-1/2/3 still trail both the AR parents and LLaDA, especially when NFE is below output length. This is also noticeable qualitatively when using the models through the demo interface. Extending Llama training from 25,000 to 50,000 updates yields a marginal average factual-accuracy gain, with improvements on some tasks offset by declines on others. While validation loss still decreases almost monotonically during the longer run, these downstream results suggest that longer training alone may be insufficient; useful ablations would include larger effective batches, alternative denoising targets, and adapters in key/output projections, MLPs, or normalization layers.

Even though we used perplexity and Distinct-1/2/3 as proxies for fluency, they are not sufficient to reliably predict human preference. For example, an external AR perplexity scoring model can favor repetitive high-probability text. While Distinct-$n$ captures this repetition, these metrics still do not measure factuality, grammar, or coherence \citep{gong2025diffullama,turok2026duel}. We therefore combine perplexity, Distinct-1/2/3, task accuracy, and demos for qualitative inspection of generated samples, rather than treating any one metric as definitive.

In the bidirectionality test, every converted model changed its prediction on every item when the later clue was added, demonstrating bidirectional attention across all four families. Some models did not always copy the token correctly, indicating a failure to use the clue rather than an absence of information flow. The public demos enable quick experimentation with NFE, output length, remasking, and blockwise generation, while exposing grammaticality, coherence, and denoising errors that aggregate metrics cannot capture.

LoRA keeps all original parameters frozen during conversion, preventing their direct overwriting and potentially reducing catastrophic forgetting \citep{luo2023catastrophic}. We adopt \(r=1024\) as a fixed transfer setting rather than claiming that it is minimal or optimal; systematically characterizing the rank–quality–compute trade-off remains future work. However, prior diffusion adaptation work also favored higher ranks over smaller ones \citep{kuiper2025lad}.

It should be noted that the high-rank adapters can subsequently be merged into the base projections. This removes their temporary parameter and inference overhead while restoring the original model size with broadly similar behavior, as shown by our experiments (Appendix~\ref{app:merge}). Future work could study whether merged models could subsequently initialize partial or full diffusion fine-tuning; establishing diffusion generation before exposing the original weights to optimization may make these updates less disruptive.

Finally, although one conversion takes only a few hours, development and evaluation across models, configurations, ablations, and validation runs required hundreds of GPU-hours. Nevertheless, BYOD reuses an AR checkpoint and processes orders of magnitude fewer token positions than the full-model systems in Table~\ref{tab:compute}, enabling repeatable cross-architecture experiments without repeated pretraining. This does not erase the original pretraining cost or establish a proportional reduction in emissions.

\section{Limitations}
The main limitation is that efficient conversion does not yet produce efficient generation, as output quality is not yet on par with the AR parents or LLaDA, especially when NFE is below output length. We also measure inference budget in network evaluations rather than wall-clock latency. Fewer sequential evaluations do not guarantee a speedup because each diffusion step processes the full answer block. Our efficiency claim therefore concerns training the conversion, not measured inference acceleration.

The cross-backbone comparison is not a controlled architecture study. The parents differ in size, tokenizer, pretraining, alignment, and attention design. Fixing LoRA rank also gives each backbone a different absolute adapter budget. Gemma's leading result therefore cannot be attributed separately to architecture, parent quality, model size, or trainable parameter count.

Finally, the automatic generation metrics cover only part of text quality. Phi-4 perplexity reflects the preferences of an external AR model, while Distinct-$n$ detects local repetition but not factuality, grammar, or coherence. Published results from other DLMs use different protocols; only LLaDA was rerun in our harness. We also did not measure energy use or carbon intensity directly, so claims on cost or resource efficiency should only be taken as indicative of relative scale rather than a measured (environmental) footprint.

\section{Conclusion}

\method{} converts instruction-tuned AR checkpoints across four families into bidirectional masked generators in a few hours on one Google Colab GPU while retaining substantial parent-model knowledge. Fluency and diversity nevertheless trail the AR parents and LLaDA, especially when NFE is below output length. A promising next step is partial or full diffusion fine-tuning initialized from the merged model, testing whether quality can improve without eroding retained capabilities. BYOD thereby enables rapid, controlled diffusion experiments across established model families with minimal compute infrastructure.

\section*{Reproducibility Statement}

The standalone implementation is available at \url{https://github.com/RuurdKuiper/BYOD}, with a Colab training notebook, revision-pinned training presets, inference and adapter merging, and the benchmark evaluation harness. Full-precision ZeroGPU demos are provided for all four conversions. The main text specifies the objective, trainable modules, data volume, optimization, generation, benchmarks, and scoring. The repository documents release scope and configuration provenance; a new training run does not guarantee bitwise reproduction of the reported checkpoints or scores.

\section*{Ethics Statement}

This study uses existing language models and public benchmark or instruction datasets and does not involve human subjects. Converted models inherit factual errors, biases, unsafe behaviors, licensing constraints, and memorization risks from their parent checkpoints and training data. We report aggregate task performance rather than claiming safety or deployment readiness; release artifacts should preserve upstream licenses and document the data provenance and intended research use.

\section*{AI Use Statement}

Generative AI tools were used to assist software development and debugging, creation of analysis scripts, literature discovery and summarization, creation of the open-ended question dataset, and drafting and editing this manuscript. The authors manually reviewed AI-assisted code and prose, and verified literature claims and citations against primary sources. No benchmark value was entered solely from an unsupported model response. The authors take responsibility for the final content, claims, code, and artifacts.

\bibliography{references}
\bibliographystyle{byod}

\appendix
\section{Cost and operational-emissions estimation}
\label{app:cost}

Table~\ref{tab:compute} compares the cost of training a model that produces diffusion behavior. It should thus be taken into account that some of these models build on an AR parent, whereas others are trained from scratch. It also excludes exploratory runs, failed jobs, dataset construction, evaluation, storage, and inference. The estimates must therefore not be interpreted as total project or lifecycle footprints.

For BYOD we use the observed 3.2 GPU-hours for the conversion of a Llama 3.1 8B model for diffusive generation. For a prior work that reports GPU-hours, we retain that value. Where a paper reports model size but not time, we approximate training compute as $C\approx6ND$ FLOPs \citep{hoffmann2022chinchilla}, where $C$ is total training compute, $N$ is the number of model parameters, and $D$ is the number of processed training-token positions. We then calculate
\begin{equation}
G_{\mathrm{est}}=\frac{6ND}{0.40\,(989\times10^{12})\,3600},
\end{equation}
where $G_{\mathrm{est}}$ is aggregate estimated GPU-hours, 0.40 is the assumed model FLOPs utilization (MFU), $989\times10^{12}$ is the NVIDIA H100's dense BF16 Tensor Core throughput in FLOP/s \citep{nvidiaH100}, and 3600 converts seconds to hours. The 40\% MFU assumption is consistent with the 39.0--45.0\% BF16 MFUs reported by FP8-LM for GPT-style 7B, 13B, and 175B training on H100 GPUs \citep{peng2023fp8lm}. Actual utilization varies considerably with architecture, sequence length, parallelization, communication, activation recomputation, and implementation efficiency. The assumption is therefore a standardized order-of-magnitude convention rather than a reconstruction of any authors' cluster. It yields the daggered values in Table~\ref{tab:compute}. The Dream and iLLaDA estimates cover their main 580B- and 12T-token stages but omit the smaller instruction-tuning additions, so they are lower bounds.

Table~\ref{tab:compute-calibration} calibrates the same calculation against studies that report both a usable token count and elapsed accelerator-hours \citep{nie2025llada,wu2025fastdllmv2}. For LLaDA, it predicts approximately 77,519 GPU-hours, compared with 130,000 reported H800 GPU-hours, corresponding to an implied utilization of 23.9\%. Thus 40\% can be optimistic about achieved utilization and correspondingly conservative with respect to estimated accelerator-hours, rental cost, and emissions. Fast-dLLM v2 displays a still larger gap. Its specialized training concatenates clean and noised views and uses a nonstandard attention pattern, illustrating that an ``implied MFU'' from $6ND$ is a calibration diagnostic, not a measured hardware-utilization statistic.

\begin{table}[!htbp]
\caption{Calibration of the $6ND$ estimator against reported accelerator-hours. Predictions use 40\% MFU and the stated accelerator's dense-BF16 peak (989 TFLOP/s H100-class proxy for H800; 312 TFLOP/s for A100) \citep{nvidiaH100,nvidiaA100}. ``Implied MFU'' is the utilization that would reconcile $6ND$ with reported time; it is not a measured MFU.}
\label{tab:compute-calibration}
\centering
\small
\begin{tabular}{lrrrrr}
\toprule
System & Parameters & Tokens & Pred. GPU-h & Reported GPU-h & Implied MFU \\
\midrule
LLaDA & 8B & 2.30T & 77,519 & 130,000 & 23.9\% \\
Fast-dLLM v2 & 7B & 1.31B & 123 & 768 & 6.4\% \\
\bottomrule
\end{tabular}
\end{table}

We convert BYOD time to a rental-equivalent cost using the September 2026 AWS US-East on-demand price of \$3.36 per hour for a one-GPU \texttt{g7e.2xlarge}, obtained from the AWS Price List. AWS specifies that this instance contains the same 96-GB NVIDIA RTX PRO 6000 Blackwell Server Edition GPU used in our Colab run \citep{awsPriceListG7e2026,awsG7e2026}. For other systems we use one consistent current major-cloud proxy: \$5.191 per accelerator-hour for an AWS \texttt{p5.4xlarge} H100 Capacity Block in US East \citep{awsCapacity2026}. Thus these figures answer ``what would the stated or estimated accelerator-hours cost at a public contemporary rate?'', not what the original authors paid. They exclude storage, egress, idle reservation time, taxes, and discounts.

Operational emissions are estimated as energy use multiplied by the data center's power usage effectiveness (PUE) and the grid carbon intensity \citep{lannelongue2021green,patterson2021carbon}:
\begin{equation}
\widehat{E}_{\mathrm{CO_2e}}=G\,P_{\mathrm{GPU}}\,\mathrm{PUE}\,I_{\mathrm{grid}}.
\end{equation}
Here $\widehat{E}_{\mathrm{CO_2e}}$ is estimated operational emissions in kg CO$_2$e, $G$ is aggregate GPU-hours, $P_{\mathrm{GPU}}$ is assumed per-GPU power draw in kW, PUE is the dimensionless data-center power usage effectiveness, and $I_{\mathrm{grid}}$ is grid carbon intensity in kg CO$_2$e/kWh. We use 0.60 kW for RTX PRO 6000 runs, matching NVIDIA's stated maximum board power \citep{nvidiaRTXPro6000}, and a 0.70-kW H100-class proxy for all other entries except the reported A100 run, for which we use 0.40 kW \citep{nvidiaH100,nvidiaA100}. We assume PUE 1.09, Google's reported fleet-wide value \citep{googlePUE2025}, and the IEA's 2024 global-average generation intensity of 0.445 kg CO$_2$e/kWh \citep{ieaElectricity2025}. For example, one BYOD stage gives $3.2\times0.60\times1.09\times0.445=0.93$ kg CO$_2$e. These are location-independent GPU-only operating scenarios: they assume full rated GPU power, omit host CPU/memory/network energy and embodied hardware emissions, and do not account for renewable-energy procurement. Direct power telemetry and the datacenter's time- and location-specific marginal carbon intensity would be required for a measured footprint.

\section{Adapter-merging ablation}
\label{app:merge}

High-rank LoRA increases the parameter count only while the adapter remains separate. To test whether removing this overhead changes behavior, we evaluate the same BYOD-Llama-25k checkpoint with the adapter loaded and after its weights are folded into the base projections. Tables~\ref{tab:merge-open-generation} and~\ref{tab:merge-benchmarks} report open-generation and factual results, respectively.

\begin{table}[!htbp]
\caption{Open-generation comparison of the adapter-loaded and merged Llama checkpoints. PPL is token-weighted; Distinct-$n$ is averaged over answers and reported as a percentage. Both use 128 generated tokens; D-1/2/3 use 32 NFE.}
\label{tab:merge-open-generation}
\centering
\scriptsize
\begin{tabular}{lrrrrrr}
\toprule
Representation & Parameters & PPL$_{32}\downarrow$ & PPL$_{128}\downarrow$ & D-1\%$\uparrow$ & D-2\%$\uparrow$ & D-3\%$\uparrow$ \\
\midrule
Adapter loaded & 8.466B & 19.381 & 5.081 & 49.5 & 78.8 & 89.9 \\
Merged standalone & 8.030B & 19.283 & 4.764 & 49.8 & 80.7 & 91.8 \\
\bottomrule
\end{tabular}
\end{table}

\begin{table}[!htbp]
\caption{Factual benchmark accuracy (\%) before and after merging in the matched harness; the mean is unweighted across the nine tasks.}
\label{tab:merge-benchmarks}
\centering
\small
\begin{tabular}{lrr@{\qquad}lrr}
\toprule
Task & Adapter & Merged & Task & Adapter & Merged \\
\midrule
ARC-C      & 83.2 & 84.0 & MATH     & 25.6 & 24.8 \\
GPQA       & 29.6 & 30.4 & MBPP     & 43.2 & 41.6 \\
GSM8K      & 65.6 & 63.2 & MMLU     & 59.2 & 60.8 \\
HellaSwag  & 70.4 & 69.6 & MMLU-Pro & 30.4 & 30.4 \\
HumanEval  & 31.2 & 29.6 & \textbf{Mean} & \textbf{48.7} & \textbf{48.3} \\
\bottomrule
\end{tabular}
\end{table}

Merging is an algebraic reparameterization rather than an additional training stage: each base projection $W$ is replaced by $W+(\alpha/r)BA$, where $A$ and $B$ are the trained rank-$r$ LoRA matrices and $\alpha$ is their scaling parameter. Subject to floating-point rounding, evaluating the folded projection is equivalent to evaluating the base and LoRA branches separately. Open-generation metrics are slightly better after merging, while factual results vary modestly by task and the macro mean decreases by only 0.4 points. These reruns therefore support preservation of overall behavior, but are not precise enough to attribute task-level changes to merging; a logit-level deterministic parity test would isolate numerical merge error from generation and benchmark variance.

\section{Complete training configuration}
\label{app:training-settings}

Table~\ref{tab:training-settings} records the common resolved configuration of the four primary conversions, including optimization, precision, checkpointing, and validation. We use AdamW with BF16 master weights and Transformer Engine HYBRID FP8 execution. The implementation pins the one-million-example training mixture \href{https://huggingface.co/datasets/Ruurd/LAD-training-1m-256}{Ruurd/LAD-training-1m-256} at revision \nolinkurl{6041513b684d6ca627879ffdb547b8629bb8b8de}.

\begin{table}[!htbp]
\caption{Full common training configuration. Model-specific parent checkpoints, mask strings, and parameter counts appear in Table~\ref{tab:models}.}
\label{tab:training-settings}
\centering
\small
\resizebox{\textwidth}{!}{%
\begin{tabular}{ll@{\qquad}ll}
\toprule
Setting & Value & Setting & Value \\
\midrule
Training examples & 400,000 & Optimizer updates & 25,000 \\
Maximum sequence length & 256 & Dynamic padding multiple & Llama: 8; others: 16 \\
Batch size & 16/16 & Gradient accumulation & 1 \\
Optimizer & AdamW & Peak learning rate & $10^{-5}$ \\
Schedule & cosine & Warmup & 100 updates \\
Weight decay & 0.01 & Maximum gradient norm & 0.5 \\
Training precision & FP8 HYBRID & Master/evaluation precision & BF16 \\
Gradient checkpointing & disabled & Selected-logit optimization & enabled \\
Corruption & IID mask-only & Noise time & $t\sim\mathcal{U}[0.001,1]$ \\
Mask-eligible region & answer + EOS padding & Loss region & corrupted answer + EOS padding \\
Answer/padding loss weights & 0.9/0.1 & Genuine answer EOS & included \\
LoRA targets & query/value projections & LoRA rank/scaling/dropout & 1024/1024/0 \\
Trainable normalization & disabled & Quantization & none \\
Validation & 200 examples/250 updates & Checkpoint interval & 1,000 updates \\
Data-loader workers & 4 & Prefetch factor & 4 \\
\bottomrule
\end{tabular}}
\end{table}

BYOD-Llama-50k keeps the same batch size and objective but processes 800,000 examples over 50,000 updates, validates every 500 updates, and uses a linear learning-rate decay from $10^{-5}$ to $10^{-6}$. This run supplies the BYOD-Llama-50k results and the diagnostics in Figure~\ref{fig:llama-training-diagnostics}.

\section{Complete inference configuration}
\label{app:inference-settings}

Table~\ref{tab:diffusion-inference} separates task-specific diffusion budgets from the training configuration and gives the evaluation sample count in one place. The fixed 100-question open-ended prompt set is \href{https://github.com/RuurdKuiper/BYOD/blob/main/src/diffusion_lm/generation_prompts.txt}{publicly available}. Unless overridden in the table, diffusion evaluation uses the official LLaDA-style fixed-transfer sampler, greedy predictions (temperature 0), low-confidence remasking, classifier-free guidance scale 0, one full-answer block, and no repetition penalty. Delayed EOS/EOT means that a predicted terminator is assigned the lowest retention confidence and may therefore be predicted but is accepted late. The open-ended diagnostic uses temperature 0.7.

\begin{table}[!htbp]
\caption{Task-specific masked-diffusion inference settings in the validation notebook. NFE is the number of denoising network evaluations.}
\label{tab:diffusion-inference}
\centering
\small
\resizebox{\textwidth}{!}{%
\begin{tabular}{lrrrrl}
\toprule
Task & Output tokens & NFE & Block length & Temperature & Terminator handling\\
\midrule
Open-ended & 128 & 128/64/32 & 128 & 0.7 & delayed retention\\
MMLU & 64 & 64 & 64 & 0 & ordinary\\
MMLU-Pro & 128 & 128 & 128 & 0 & delayed retention\\
HellaSwag & 64 & 64 & 64 & 0 & ordinary\\
ARC-Challenge & 64 & 64 & 64 & 0 & ordinary\\
GSM8K & 256 & 256 & 256 & 0 & delayed retention\\
MATH-500 & 256 & 256 & 256 & 0 & delayed retention\\
GPQA-Main & 128 & 128 & 128 & 0 & delayed retention\\
HumanEval & 256 & 256 & 256 & 0 & delayed retention\\
MBPP & 128 & 128 & 128 & 0 & delayed retention\\
Bidirectional attention & 1 masked word & 1 & \na & 0 & \na\\
\bottomrule
\end{tabular}
}
\end{table}

\begin{table}[!htbp]
\caption{Maximum new tokens for paired greedy AR evaluation.}
\label{tab:ar-inference}
\centering
\small
\begin{tabular}{lr@{\qquad}lr}
\toprule
Task & Tokens & Task & Tokens \\
\midrule
Open-ended & 128 & MMLU & 64 \\
MMLU-Pro & 2,048 & HellaSwag & 64 \\
ARC-Challenge & 64 & GSM8K & 256 \\
MATH-500 & 256 & GPQA-Main & 128 \\
HumanEval & 256 & MBPP & 128 \\
\bottomrule
\end{tabular}
\end{table}

Table~\ref{tab:ar-inference} reports the corresponding maximum-new-token budgets for greedy AR evaluation. These are listed separately because an AR token budget is not an NFE budget and should not be read as computational equivalence. For computational tractability, we evaluate all models on the same fixed subset of 125 examples from each benchmark, selected once using a fixed seed. Thus, comparisons within our harness are paired at the example level. This enables broad, paired comparison across multiple model families, training checkpoints, and inference settings, while avoiding the substantially greater computational cost that would result from full-benchmark evaluation on top of the extensive preliminary hyperparameter and configuration testing required to establish a stable conversion recipe.

\section{Qualitative code-error analysis}
\label{app:code-errors}

We reanalyzed the saved BYOD-Llama-50k and Llama-AR outputs underlying Table~\ref{tab:accuracy}. After applying the benchmark harness's ordinary Markdown-code extraction, we passed each resulting program to Python's \texttt{ast.parse} before executing tests. Table~\ref{tab:syntax-errors} therefore counts only syntax-invalid programs; syntactically valid but incorrect programs are not included.

\begin{table}[!htbp]
\caption{Parser failures on paired code-generation outputs.}
\label{tab:syntax-errors}
\centering
\small
\begin{tabular}{llrr}
\toprule
Model & Benchmark & Invalid / 125 & Rate \\
\midrule
BYOD-Llama-50k & HumanEval & 35 & 28.0\% \\
Llama AR & HumanEval & 0 & 0.0\% \\
BYOD-Llama-50k & MBPP & 11 & 8.8\% \\
Llama AR & MBPP & 0 & 0.0\% \\
\bottomrule
\end{tabular}
\end{table}

Table~\ref{tab:syntax-examples} shows verbatim fragments from four parser-invalid BYOD-Llama-50k generations. The intended program structure is recognizable, but a local omission, duplication, or delimiter error makes each output unexecutable and therefore a pass@1 failure. A local syntax repair would not necessarily make the complete program semantically correct. These examples and rates establish that syntax errors materially contribute to the observed code gap; they are consistent with local-consistency failures under parallel prediction, but do not by themselves establish the decoding mechanism as their cause.

\begin{table}[!htbp]
\caption{Representative local syntax errors in saved BYOD-Llama-50k outputs. Fragments are reproduced verbatim after the same code extraction used for evaluation.}
\label{tab:syntax-examples}
\centering
\small
\begin{tabular}{p{0.14\textwidth}p{0.50\textwidth}p{0.25\textwidth}}
\toprule
Item & Generated fragment & Parser-level failure \\
\midrule
HumanEval/10 & \texttt{defdef make\_palindrome(string: str) -> str:} & Duplicated \texttt{def} keyword. \\
HumanEval/2 & \texttt{return number - int(number} & Missing closing parenthesis. \\
MBPP/25 & \texttt{return [t for t in tuples if all(x \% k == 0 for x in t]} & Square bracket closes an open parenthesis. \\
MBPP/79 & \texttt{return set(num for for sublist in list\_of\_lists)} & Duplicated \texttt{for} keyword. \\
\bottomrule
\end{tabular}
\end{table}

\section{Complete open-generation results}
\label{app:open-generation-results}

\begin{figure}[!htbp]
\centering
\includegraphics[width=\textwidth]{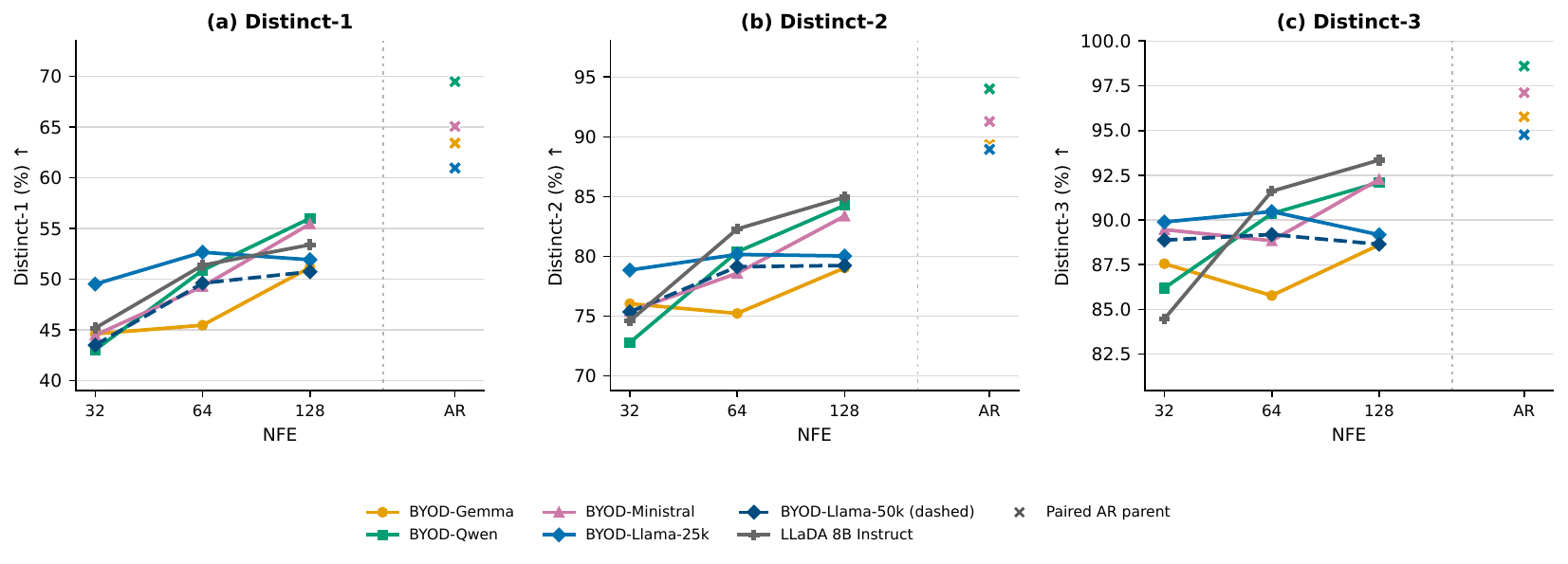}
\caption{Sliding model-token Distinct-1/2/3 for the final models across diffusion inference budgets. Crosses show the paired AR parents; BYOD-Llama-25k and BYOD-Llama-50k share the same parent. Higher-order metrics show the same broad diversity pattern while compressing differences because longer $n$-grams are less likely to repeat exactly.}
\label{fig:final-distinct-metrics}
\end{figure}

\begin{figure}[!htbp]
\centering
\includegraphics[width=0.92\textwidth]{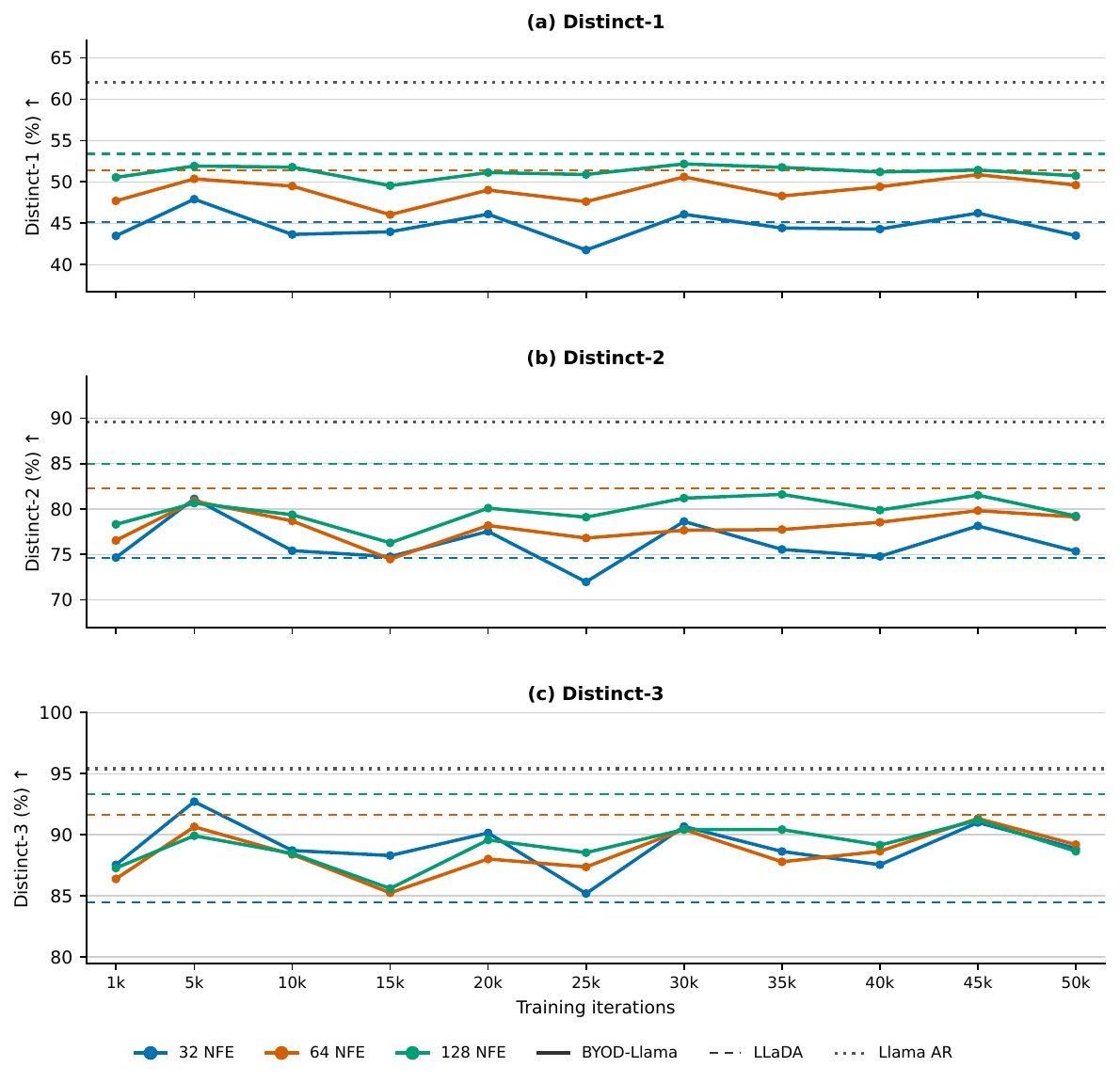}
\caption{Sliding model-token Distinct-1/2/3 throughout BYOD-Llama-50k training. Solid colored lines are BYOD-Llama checkpoints at each NFE budget; matching dashed lines are fixed LLaDA references, and the dotted gray line is Llama AR. All three metrics fluctuate across checkpoints without a strong monotonic trend.}
\label{fig:llama-distinct-progression}
\end{figure}

Table~\ref{tab:ppl} gives the exact perplexity and Distinct-1/2/3 values for the final checkpoints. Figure~\ref{fig:final-distinct-metrics} compares the diversity metrics across inference budgets, while Figure~\ref{fig:llama-distinct-progression} shows their development during BYOD-Llama-50k training.

\begin{table}[!htbp]
\caption{Open-ended generation on a 128-token canvas. PPL$_k$ is token-weighted Phi-4 perplexity; D-$n$ columns report mean per-answer sliding model-token Distinct-$n$ at $k$ NFE. Lower PPL and higher Distinct-$n$ are better. AR parents generate autoregressively and therefore have no diffusion-NFE setting.}
\label{tab:ppl}
\centering
\scriptsize
\par\noindent\textbf{(a) Perplexity and Distinct-1.}
\vspace{0.3em}
\resizebox{\textwidth}{!}{%
\begin{tabular}{lrrrrrrrr}
\toprule
& \multicolumn{6}{c}{Diffusion} & \multicolumn{2}{c}{AR parent} \\
\cmidrule(lr){2-7}\cmidrule(lr){8-9}
Model/checkpoint & PPL$_{32}\downarrow$ & D-1$_{32}$\%$\uparrow$ & PPL$_{64}\downarrow$ & D-1$_{64}$\%$\uparrow$ & PPL$_{128}\downarrow$ & D-1$_{128}$\%$\uparrow$ & PPL$\downarrow$ & D-1\%$\uparrow$ \\
\midrule
BYOD-Gemma & 14.56 & 44.6 & 8.27 & 45.5 & 4.33 & 51.2 & 3.01 & 63.4 \\
BYOD-Qwen & 20.25 & 43.0 & 11.12 & 50.9 & 6.21 & 56.0 & 2.23 & 69.5 \\
BYOD-Ministral & 15.45 & 44.5 & 9.42 & 49.3 & 5.08 & 55.5 & 2.31 & 65.1 \\
BYOD-Llama-25k & 19.38 & 49.5 & 10.50 & 52.7 & 5.08 & 51.9 & 2.49 & 61.0 \\
BYOD-Llama-50k & 17.00 & 43.5 & 9.53 & 49.6 & 4.83 & 50.7 & 2.49 & 61.0 \\
LLaDA 8B Instruct & 5.33 & 45.2 & 4.07 & 51.4 & 2.79 & 53.4 & \na & \na \\
\bottomrule
\end{tabular}}
\input{generated_distinct_results.tex}
\end{table}

\section{Bidirectional attention test}
\label{app:bidirectional-infilling}

This deterministic test uses ordinary single-token target words. Each partial response has the form ``Answer:[MASK]. COPY: apple.'', with a different target in each example. The target is absent from the instruction and left context and appears explicitly after the mask. The converted model predicts the masked position by greedy argmax in one forward pass; no iterative sampling or generation settings are involved.

We score the same masked position twice: once with the later copy clue and once after deleting that clue. We report exact match for both conditions, their difference, and whether adding the clue changes the greedy prediction at all. Under causal attention, changing or deleting tokens to the right cannot affect logits at an earlier position, so both the exact-match gain and prediction-change rate must be zero. The latter separates use of right context from the stricter ability to follow the exact-copy instruction. This is a functional attention check, not a benchmark of reasoning or general infilling quality. Table~\ref{tab:bidirectional-infilling} reports the results for both conditions.

\begin{table}[!htbp]
\caption{Bidirectional attention test results. ``With clue'' predicts the earlier mask while the word is visible later; ``without clue'' removes that later word. Gain is the exact-match difference.}
\label{tab:bidirectional-infilling}
\centering
\scriptsize
\resizebox{\columnwidth}{!}{%
\begin{tabular}{lcccc}
\toprule
Model & With clue EM & Without clue EM & EM gain & Prediction changed \\
\midrule
BYOD-Gemma & 100\% & 0\% & +100 pp & 100\% \\
BYOD-Qwen & 20\% & 0\% & +20 pp & 100\% \\
BYOD-Ministral & 100\% & 0\% & +100 pp & 100\% \\
BYOD-Llama-25k & 90\% & 0\% & +90 pp & 100\% \\
BYOD-Llama-50k & 70\% & 0\% & +70 pp & 100\% \\
LLaDA 8B Instruct & 100\% & 0\% & +100 pp & 100\% \\
\bottomrule
\end{tabular}}
\end{table}

\section{Protocol-separated literature results}
\label{app:published-results}

Table~\ref{tab:published} places our primary Llama conversion beside values reported by other diffusion-model papers \citep{nie2025llada,ye2025dream,nie2026illada,fu2025efficientdlm}. The BYOD column uses our common harness, whereas every other column uses its source paper's evaluation protocol. Differences in prompts, subsets, generation budgets, answer extraction, and execution environments prevent direct ranking; the table is included only to provide scale and task coverage. NR means that the source did not report a value for that task.

\begin{table}[!htbp]
\caption{Our Llama result and protocol-separated published diffusion-model accuracies. Bold marks the largest reported value in each row; it does not imply a controlled comparison. No cross-task averages are shown because task coverage differs by model.}
\label{tab:published}
\centering
\scriptsize
\resizebox{\columnwidth}{!}{%
\begin{tabular}{lccccc}
\toprule
Task & BYOD-Llama (ours) & LLaDA & Dream-Inst. & iLLaDA-Inst. & Efficient-DLM 8B \\
\midrule
ARC-C     & .832 & \textbf{.885} & NR & NR & .619 \\
GPQA      & .296 & \textbf{.333} & .330 & NR & NR \\
GSM8K     & .656 & .694 & .810 & \textbf{.890} & .883 \\
HellaSwag & .704 & \textbf{.746} & NR & NR & .725 \\
HumanEval & .312 & .494 & .555 & .659 & \textbf{.683} \\
MATH      & .256 & .319 & .392 & \textbf{.567} & .501 \\
MBPP      & .432 & .410 & .588 & .580 & \textbf{.610} \\
MMLU      & .592 & .655 & .670 & .716 & \textbf{.772} \\
MMLU-Pro  & .304 & .370 & .433 & \textbf{.523} & NR \\
\bottomrule
\end{tabular}}
\end{table}

\section{Interactive denoising trajectories}
\label{app:demo-trajectories}

\begin{figure}[!htbp]
\centering
\includegraphics[width=0.85\textwidth]{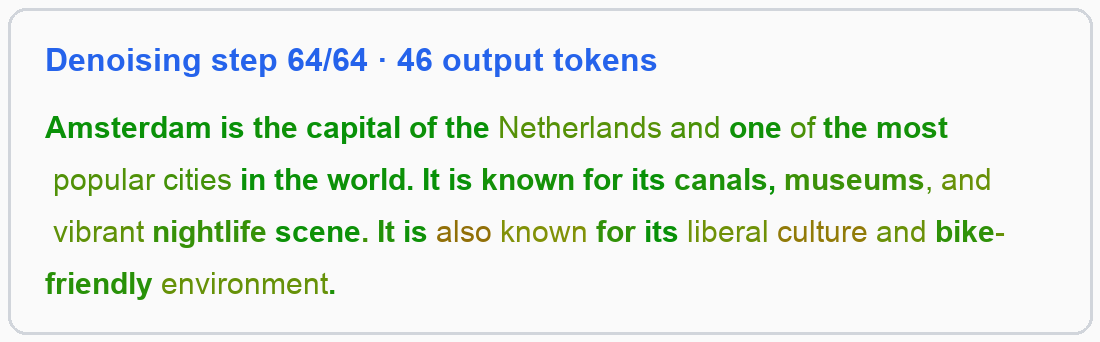}
\\[-2pt]\small (a) 64 NFE and 64 tokens.

\vspace{3pt}
\includegraphics[width=0.85\textwidth]{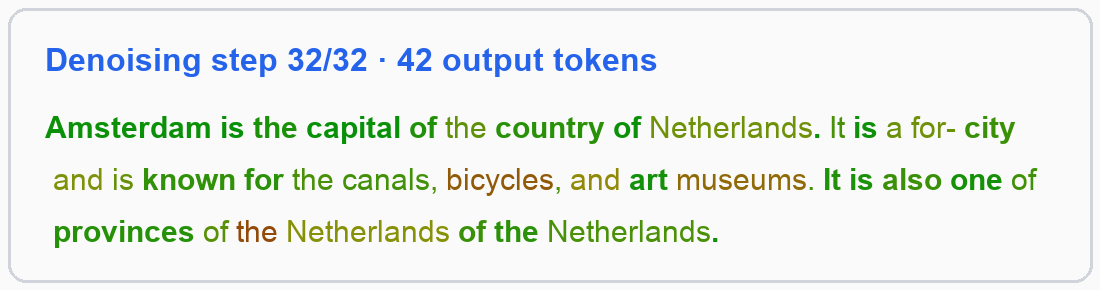}
\\[-2pt]\small (b) 32 NFE and 64 tokens.

\vspace{3pt}
\includegraphics[width=0.85\textwidth]{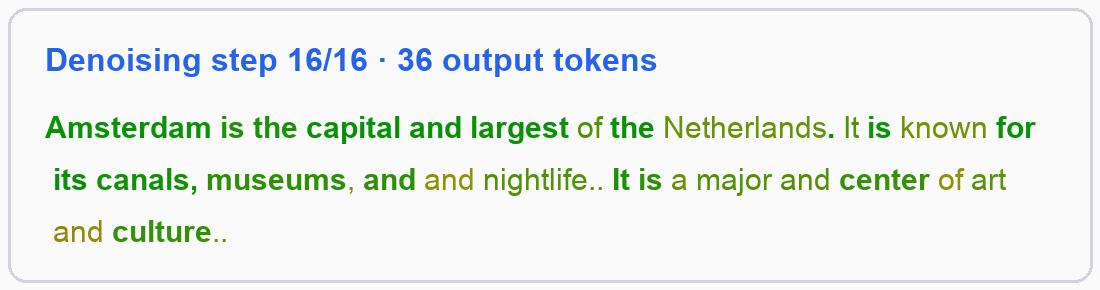}
\\[-2pt]\small (c) 16 NFE and 64 tokens.
\caption{Confidence-guided BYOD-Gemma-2-9B trajectories at decreasing inference budgets on a 64-token canvas. Red-to-green coloring records each retained token's sampling probability.}
\label{fig:demo-confidence-trajectories}
\end{figure}

\begin{figure}[!htbp]
\centering
\includegraphics[width=0.85\textwidth]{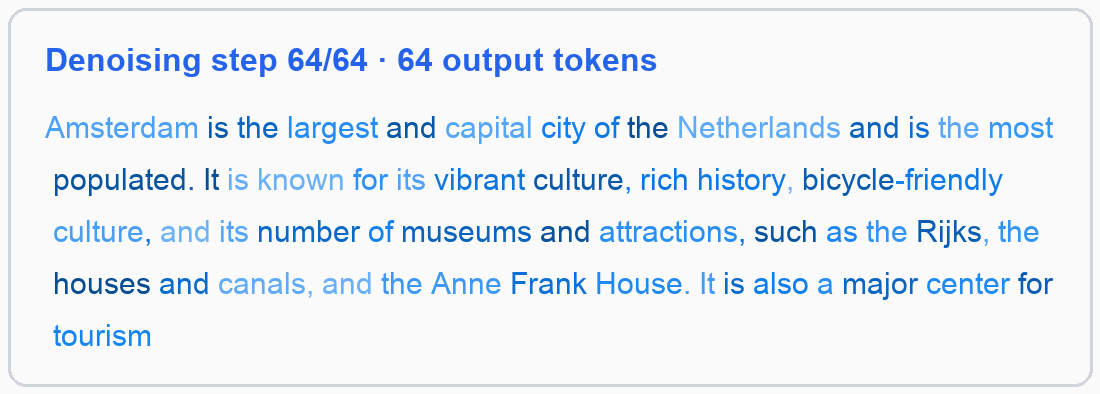}
\\[-2pt]\small (a) Random remasking; prediction-iteration coloring.

\vspace{3pt}
\includegraphics[width=0.85\textwidth]{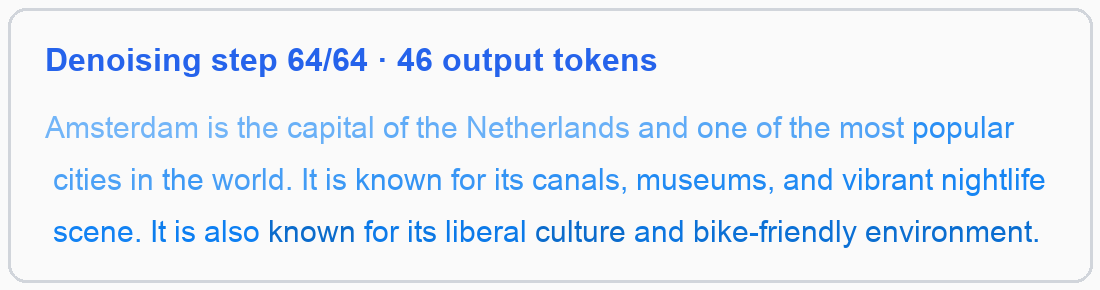}
\\[-2pt]\small (b) Confidence-guided remasking; prediction-iteration coloring.
\caption{Remasking comparison for BYOD-Gemma-2-9B at 64 NFE on a 64-token canvas. Blue intensity records the iteration at which each retained token was predicted.}
\label{fig:demo-remasking-trajectories}
\end{figure}

Figures~\ref{fig:demo-confidence-trajectories} and~\ref{fig:demo-remasking-trajectories} illustrate how the public BYOD-Gemma demo exposes the parallel refinement process under random and confidence-guided remasking. All panels use the displayed Amsterdam prompt, the helpful-assistant system prompt, temperature 0.7, top-$k$ 3, initial noise fraction 1.0, delayed EOS/EOT retention, permanent locked retention, and no visualization delay. Early stopping is disabled and block length equals the new-token canvas, ensuring that every panel displays its complete denoising budget. The demo randomizes the hidden sampling seed for every request; we resample the 64-NFE conditions to show readable positive examples, so these trajectories illustrate behavior rather than constitute a controlled quality comparison.

\end{document}

%% file: math_commands.tex
\usepackage{amsmath,amsfonts,bm}

\def\eqref#1{equation~\ref{#1}}

\def\1{\bm{1}}

\DeclareMathAlphabet{\mathsfit}{\encodingdefault}{\sfdefault}{m}{sl}
\SetMathAlphabet{\mathsfit}{bold}{\encodingdefault}{\sfdefault}{bx}{n}



%% file: authors.tex
\author{Ruurd J. A. Kuiper$^{1}$ \qquad Robert A. Bagheri$^{2}$\\
$^{1}$Julius Center for Health Sciences and Primary Care\\
University Medical Center Utrecht\\
$^{2}$Department of Methodology and Statistics\\
Utrecht University, The Netherlands\\
\texttt{r.j.a.kuiper@umcutrecht.nl}
}

%% file: generated_distinct_results.tex
\vspace{0.8em}
\par\noindent\textbf{(b) Higher-order diversity.} D-2 and D-3 are mean per-answer sliding model-token Distinct-2 and Distinct-3, in percent.
\vspace{0.3em}
\resizebox{\textwidth}{!}{%
\begin{tabular}{lrrrrrrrr}
\toprule
& \multicolumn{6}{c}{Diffusion} & \multicolumn{2}{c}{AR parent} \\
\cmidrule(lr){2-7}\cmidrule(lr){8-9}
Model/checkpoint & D-2$_{32}$ & D-3$_{32}$ & D-2$_{64}$ & D-3$_{64}$ & D-2$_{128}$ & D-3$_{128}$ & D-2 & D-3 \\
\midrule
BYOD-Gemma & 76.0 & 87.6 & 75.2 & 85.8 & 79.0 & 88.6 & 89.3 & 95.8 \\
BYOD-Qwen & 72.8 & 86.2 & 80.4 & 90.4 & 84.3 & 92.1 & 94.0 & 98.6 \\
BYOD-Ministral & 75.4 & 89.5 & 78.6 & 88.8 & 83.4 & 92.3 & 91.3 & 97.1 \\
BYOD-Llama-25k & 78.8 & 89.9 & 80.2 & 90.5 & 80.0 & 89.2 & 89.0 & 94.8 \\
BYOD-Llama-50k & 75.3 & 88.9 & 79.1 & 89.2 & 79.2 & 88.6 & 89.0 & 94.8 \\
LLaDA 8B Instruct & 74.6 & 84.5 & 82.3 & 91.6 & 84.9 & 93.3 & \na & \na \\
\bottomrule
\end{tabular}}

%% file: paper.bbl
\begin{thebibliography}{52}
\providecommand{\natexlab}[1]{#1}
\providecommand{\url}[1]{\texttt{#1}}
\expandafter\ifx\csname urlstyle\endcsname\relax
  \providecommand{\doi}[1]{doi: #1}\else
  \providecommand{\doi}{doi: \begingroup \urlstyle{rm}\Url}\fi

\bibitem[{Amazon Web Services}(2026{\natexlab{a}})]{awsCapacity2026}
{Amazon Web Services}.
\newblock Amazon {EC2} capacity blocks for {ML} pricing.
\newblock \url{https://aws.amazon.com/ec2/capacityblocks/pricing/},
  2026{\natexlab{a}}.
\newblock Accessed 2026-09-21; reports \$5.191 per accelerator-hour for
  p5.4xlarge in US East.

\bibitem[{Amazon Web Services}(2026{\natexlab{b}})]{awsG7e2026}
{Amazon Web Services}.
\newblock Amazon {EC2 G7e} instances.
\newblock \url{https://aws.amazon.com/ec2/instance-types/g7e/},
  2026{\natexlab{b}}.
\newblock Accessed 2026-09-25; specifies one 96-GB NVIDIA RTX PRO 6000
  Blackwell Server Edition GPU for g7e.2xlarge.

\bibitem[{Amazon Web Services}(2026{\natexlab{c}})]{awsPriceListG7e2026}
{Amazon Web Services}.
\newblock {AWS Price List}: Amazon {EC2}, {US East (Ohio)}.
\newblock
  \url{https://pricing.us-east-1.amazonaws.com/offers/v1.0/aws/AmazonEC2/current/us-east-2/index.json},
  2026{\natexlab{c}}.
\newblock Accessed 2026-09-25; SKU 84MJTJ2X73HAYM9Q lists \$3.36312 per
  on-demand Linux g7e.2xlarge instance-hour.

\bibitem[Arriola et~al.(2025)Arriola, Gokaslan, Chiu, Yang, Qi, Han, Sahoo, and
  Kuleshov]{arriola2025block}
Marianne Arriola, Aaron Gokaslan, Justin~T. Chiu, Zhihan Yang, Zhixuan Qi,
  Jiaqi Han, Subham~Sekhar Sahoo, and Volodymyr Kuleshov.
\newblock Block diffusion: Interpolating between autoregressive and diffusion
  language models.
\newblock \emph{arXiv preprint arXiv:2503.09573}, 2025.

\bibitem[Austin et~al.(2021{\natexlab{a}})Austin, Johnson, Ho, Tarlow, and
  van~den Berg]{austin2021structured}
Jacob Austin, Daniel~D. Johnson, Jonathan Ho, Daniel Tarlow, and Rianne van~den
  Berg.
\newblock Structured denoising diffusion models in discrete state-spaces.
\newblock In \emph{Advances in Neural Information Processing Systems},
  2021{\natexlab{a}}.

\bibitem[Austin et~al.(2021{\natexlab{b}})Austin, Odena, Nye, Bosma,
  Michalewski, Dohan, Jiang, Cai, Terry, Le, and Sutton]{austin2021mbpp}
Jacob Austin, Augustus Odena, Maxwell Nye, Maarten Bosma, Henryk Michalewski,
  David Dohan, Ellen Jiang, Carrie Cai, Michael Terry, Quoc Le, and Charles
  Sutton.
\newblock Program synthesis with large language models.
\newblock \emph{arXiv preprint arXiv:2108.07732}, 2021{\natexlab{b}}.

\bibitem[Bie et~al.(2025)Bie, Cao, Chen, Du, Gong, Gong, Gu, Hu, Huang, Lan,
  Li, Li, Li, Li, Liu, Liu, Lu, Lu, Ma, Tan, Wei, Wen, Xing, Zhang, Zhao,
  Zheng, Zhou, Zhou, Zhou, Zhu, and Zhuang]{bie2025llada20}
Tiwei Bie, Maosong Cao, Kun Chen, Lun Du, Mingliang Gong, Zhuochen Gong, Yanmei
  Gu, Jiaqi Hu, Zenan Huang, Zhenzhong Lan, Chengxi Li, Chongxuan Li, Jianguo
  Li, Zehuan Li, Huabin Liu, Ling Liu, Guoshan Lu, Xiaocheng Lu, Yuxin Ma,
  Jianfeng Tan, Lanning Wei, Ji-Rong Wen, Yipeng Xing, Xiaolu Zhang, Junbo
  Zhao, Da~Zheng, Jun Zhou, Junlin Zhou, Zhanchao Zhou, Liwang Zhu, and Yihong
  Zhuang.
\newblock {LLaDA2.0}: Scaling up diffusion language models to 100b.
\newblock \emph{arXiv preprint arXiv:2512.15745}, 2025.

\bibitem[Chandrasegaran et~al.(2025)Chandrasegaran, Thomas, Ku, Berto, Kim,
  Brixi, Nguyen, Massaroli, and Poli]{chandrasegaran2025rnd1}
Keshigeyan Chandrasegaran, Armin~W. Thomas, Jerome Ku, Federico Berto,
  Jae~Myung Kim, Garyk Brixi, Eric Nguyen, Stefano Massaroli, and Michael Poli.
\newblock {RND1}: Simple, scalable {AR}-to-diffusion conversion.
\newblock Radical Numerics technical report, October 2025.
\newblock URL \url{https://www.radicalnumerics.ai/assets/rnd1_report.pdf}.

\bibitem[Chen et~al.(2021)]{chen2021codex}
Mark Chen et~al.
\newblock Evaluating large language models trained on code.
\newblock \emph{arXiv preprint arXiv:2107.03374}, 2021.

\bibitem[Clark et~al.(2018)Clark, Cowhey, Etzioni, Khot, Sabharwal, Schoenick,
  and Tafjord]{clark2018arc}
Peter Clark, Isaac Cowhey, Oren Etzioni, Tushar Khot, Ashish Sabharwal, Carissa
  Schoenick, and Oyvind Tafjord.
\newblock Think you have solved question answering? try {ARC}, the {AI2}
  reasoning challenge.
\newblock In \emph{arXiv preprint arXiv:1803.05457}, 2018.

\bibitem[Cobbe et~al.(2021)Cobbe, Kosaraju, Bavarian, Chen, Jun, Kaiser,
  Plappert, Tworek, Hilton, Nakano, Hesse, and Schulman]{cobbe2021gsm8k}
Karl Cobbe, Vineet Kosaraju, Mohammad Bavarian, Mark Chen, Heewoo Jun, Lukasz
  Kaiser, Matthias Plappert, Jerry Tworek, Jacob Hilton, Rei Nakano,
  Christopher Hesse, and John Schulman.
\newblock Training verifiers to solve math word problems.
\newblock \emph{arXiv preprint arXiv:2110.14168}, 2021.

\bibitem[{Crumb}(2023)]{cleaninstruct3m}
{Crumb}.
\newblock Clean-instruct-3m.
\newblock Dataset release, 2023.
\newblock URL \url{https://huggingface.co/datasets/crumb/Clean-Instruct-3M}.

\bibitem[{DiffusionGemma Team}(2026)]{diffusiongemma2026}
{DiffusionGemma Team}.
\newblock {DiffusionGemma} technical report.
\newblock \emph{arXiv preprint arXiv:2608.00146}, 2026.

\bibitem[Fu et~al.(2025)Fu, Whalen, Ye, Dong, Diao, Liu, Wu, Zhang, Xie, Han,
  Khadkevich, Kautz, Lin, and Molchanov]{fu2025efficientdlm}
Yonggan Fu, Lexington Whalen, Zhifan Ye, Xin Dong, Shizhe Diao, Jingyu Liu,
  Chengyue Wu, Hao Zhang, Enze Xie, Song Han, Maksim Khadkevich, Jan Kautz,
  Yingyan~Celine Lin, and Pavlo Molchanov.
\newblock Efficient-{DLM}: From autoregressive to diffusion language models,
  and beyond in speed.
\newblock \emph{arXiv preprint arXiv:2512.14067}, 2025.

\bibitem[Gong et~al.(2025)Gong, Agarwal, Zhang, Ye, Zheng, Li, An, Zhao, Bi,
  Han, Peng, and Kong]{gong2025diffullama}
Shansan Gong, Shivam Agarwal, Yizhe Zhang, Jiacheng Ye, Lin Zheng, Mukai Li,
  Chenxin An, Peilin Zhao, Wei Bi, Jiawei Han, Hao Peng, and Lingpeng Kong.
\newblock Scaling diffusion language models via adaptation from autoregressive
  models.
\newblock In \emph{International Conference on Learning Representations}, 2025.

\bibitem[{Google Cloud}(2025)]{googlePUE2025}
{Google Cloud}.
\newblock Building software sustainably.
\newblock
  \url{https://cloud.google.com/blog/topics/sustainability/building-software-sustainably},
  2025.
\newblock Accessed 2026-09-21; reports fleet-wide PUE of 1.09.

\bibitem[Hendrycks et~al.(2021{\natexlab{a}})Hendrycks, Burns, Basart, Zou,
  Mazeika, Song, and Steinhardt]{hendrycks2021mmlu}
Dan Hendrycks, Collin Burns, Steven Basart, Andy Zou, Mantas Mazeika, Dawn
  Song, and Jacob Steinhardt.
\newblock Measuring massive multitask language understanding.
\newblock In \emph{International Conference on Learning Representations},
  2021{\natexlab{a}}.

\bibitem[Hendrycks et~al.(2021{\natexlab{b}})Hendrycks, Burns, Kadavath, Arora,
  Basart, Tang, Song, and Steinhardt]{hendrycks2021math}
Dan Hendrycks, Collin Burns, Saurav Kadavath, Akul Arora, Steven Basart, Eric
  Tang, Dawn Song, and Jacob Steinhardt.
\newblock Measuring mathematical problem solving with the {MATH} dataset.
\newblock In \emph{Advances in Neural Information Processing Systems},
  2021{\natexlab{b}}.

\bibitem[Hersche et~al.(2026)Hersche, Menet, Tanios, and
  Rahimi]{hersche2026codila}
Michael Hersche, Nicolas Menet, Ronan Tanios, and Abbas Rahimi.
\newblock Locally coherent parallel decoding in diffusion language models.
\newblock \emph{arXiv preprint arXiv:2603.20216}, 2026.
\newblock URL \url{https://arxiv.org/abs/2603.20216}.

\bibitem[Hoffmann et~al.(2022)Hoffmann, Borgeaud, Mensch, Buchatskaya, Cai,
  Rutherford, de~Las~Casas, Hendricks, Welbl, Clark, Hennigan, Noland,
  Millican, van~den Driessche, Damoc, Guy, Osindero, Simonyan, Elsen, Vinyals,
  Rae, and Sifre]{hoffmann2022chinchilla}
Jordan Hoffmann, Sebastian Borgeaud, Arthur Mensch, Elena Buchatskaya, Trevor
  Cai, Eliza Rutherford, Diego de~Las~Casas, Lisa~Anne Hendricks, Johannes
  Welbl, Aidan Clark, Tom Hennigan, Eric Noland, Katie Millican, George van~den
  Driessche, Bogdan Damoc, Aurelia Guy, Simon Osindero, Karen Simonyan, Erich
  Elsen, Oriol Vinyals, Jack~W. Rae, and Laurent Sifre.
\newblock Training compute-optimal large language models.
\newblock In \emph{Advances in Neural Information Processing Systems},
  volume~35, pp.\  30016--30030, 2022.

\bibitem[Hoogeboom et~al.(2021)Hoogeboom, Nielsen, Jaini, Forr{\'e}, and
  Welling]{hoogeboom2021multinomial}
Emiel Hoogeboom, Didrik Nielsen, Priyank Jaini, Patrick Forr{\'e}, and Max
  Welling.
\newblock Argmax flows and multinomial diffusion: Learning categorical
  distributions.
\newblock In \emph{Advances in Neural Information Processing Systems},
  volume~34, pp.\  12454--12465, 2021.

\bibitem[Hu et~al.(2022)Hu, Shen, Wallis, Allen-Zhu, Li, Wang, Wang, and
  Chen]{hu2022lora}
Edward~J. Hu, Yelong Shen, Phillip Wallis, Zeyuan Allen-Zhu, Yuanzhi Li, Shean
  Wang, Lu~Wang, and Weizhu Chen.
\newblock {LoRA}: Low-rank adaptation of large language models.
\newblock In \emph{International Conference on Learning Representations}, 2022.

\bibitem[Huang et~al.(2024)Huang, Cheng, Liu, Hao, Song, Xu,
  et~al.]{huang2024opencoder}
Siming Huang, Tianhao Cheng, Jason~Klein Liu, Jiaran Hao, Liuyihan Song, Yang
  Xu, et~al.
\newblock Opencoder: The open cookbook for top-tier code large language models.
\newblock \emph{arXiv preprint arXiv:2411.04905}, 2024.

\bibitem[{International Energy Agency}(2025)]{ieaElectricity2025}
{International Energy Agency}.
\newblock Electricity 2025: Emissions.
\newblock \url{https://www.iea.org/reports/electricity-2025/emissions}, 2025.
\newblock Reports global electricity-generation intensity of 445 g CO2/kWh in
  2024.

\bibitem[Kuiper et~al.(2025)Kuiper, de~Groot, van Es, van Smeden, and
  Bagheri]{kuiper2025lad}
Ruurd J.~A. Kuiper, Lars de~Groot, Bram van Es, Maarten van Smeden, and Ayoub
  Bagheri.
\newblock {LAD}: {LoRA}-adapted diffusion.
\newblock In \emph{Proceedings of the 2025 Conference on Empirical Methods in
  Natural Language Processing: System Demonstrations}, pp.\  97--110, 2025.

\bibitem[Lambert et~al.(2024)Lambert, Morrison, Pyatkin, Huang, Ivison,
  Brahman, et~al.]{lambert2024tulu3}
Nathan Lambert, Jacob Morrison, Valentina Pyatkin, Shengyi Huang, Hamish
  Ivison, Faeze Brahman, et~al.
\newblock T\"ulu 3: Pushing frontiers in open language model post-training.
\newblock \emph{arXiv preprint arXiv:2411.15124}, 2024.

\bibitem[Lannelongue et~al.(2021)Lannelongue, Grealey, and
  Inouye]{lannelongue2021green}
Lo{\"i}c Lannelongue, Jason Grealey, and Michael Inouye.
\newblock Green algorithms: Quantifying the carbon footprint of computation.
\newblock \emph{Advanced Science}, 8\penalty0 (12):\penalty0 2100707, 2021.
\newblock \doi{10.1002/advs.202100707}.

\bibitem[Li et~al.(2016)Li, Galley, Brockett, Gao, and Dolan]{li2016diversity}
Jiwei Li, Michel Galley, Chris Brockett, Jianfeng Gao, and Bill Dolan.
\newblock A diversity-promoting objective function for neural conversation
  models.
\newblock In \emph{Proceedings of the 2016 Conference of the North American
  Chapter of the Association for Computational Linguistics: Human Language
  Technologies}, pp.\  110--119, 2016.
\newblock \doi{10.18653/v1/N16-1014}.

\bibitem[Luo et~al.(2023)Luo, Yang, Meng, Li, Zhou, and
  Zhang]{luo2023catastrophic}
Yun Luo, Zhen Yang, Fandong Meng, Yafu Li, Jie Zhou, and Yue Zhang.
\newblock An empirical study of catastrophic forgetting in large language
  models during continual fine-tuning.
\newblock \emph{arXiv preprint arXiv:2308.08747}, 2023.

\bibitem[Mitra et~al.(2024)Mitra, Khanpour, Rosset, and
  Awadallah]{mitra2024orcamath}
Arindam Mitra, Hamed Khanpour, Corby Rosset, and Ahmed Awadallah.
\newblock Orca-math: Unlocking the potential of {SLMs} in grade school math.
\newblock \emph{arXiv preprint arXiv:2402.14830}, 2024.

\bibitem[Nie et~al.(2025)Nie, Zhu, You, Zhang, Ou, Hu, Zhou, Lin, Wen, and
  Li]{nie2025llada}
Shen Nie, Fengqi Zhu, Zebin You, Xiaolu Zhang, Jingyang Ou, Jun Hu, Jun Zhou,
  Yankai Lin, Ji-Rong Wen, and Chongxuan Li.
\newblock Large language diffusion models.
\newblock \emph{arXiv preprint arXiv:2502.09992}, 2025.

\bibitem[Nie et~al.(2026)Nie, Min, Xu, Huang, Song, Shan, Lin, Zhao, Li, and
  Wen]{nie2026illada}
Shen Nie, Qiyang Min, Shaoxuan Xu, Zihao Huang, Yuxuan Song, Yong Shan, Yankai
  Lin, Wayne~Xin Zhao, Chongxuan Li, and Ji-Rong Wen.
\newblock Improved large language diffusion models.
\newblock \emph{arXiv preprint arXiv:2606.25331}, 2026.

\bibitem[{NVIDIA}(2021)]{nvidiaA100}
{NVIDIA}.
\newblock {NVIDIA A100 Tensor Core GPU} data sheet.
\newblock
  \url{https://www.nvidia.com/content/dam/en-zz/Solutions/Data-Center/a100/pdf/nvidia-a100-datasheet-nvidia-us-2188504-web.pdf},
  2021.
\newblock Accessed 2026-09-21; SXM maximum thermal design power 400 W.

\bibitem[{NVIDIA}(2023)]{nvidiaH100}
{NVIDIA}.
\newblock {NVIDIA H100 Tensor Core GPU}.
\newblock \url{https://www.nvidia.com/en-us/data-center/h100/}, 2023.
\newblock Accessed 2026-09-21; dense BF16 Tensor Core peak 989 TFLOP/s and
  configurable TDP up to 700 W.

\bibitem[{NVIDIA}(2025)]{nvidiaRTXPro6000}
{NVIDIA}.
\newblock {NVIDIA RTX PRO 6000 Blackwell Server Edition}.
\newblock
  \url{https://www.nvidia.com/en-us/data-center/rtx-pro-6000-blackwell-server-edition/},
  2025.
\newblock Accessed 2026-09-21; maximum power consumption up to 600 W.

\bibitem[Ou et~al.(2025)Ou, Nie, Xue, Zhu, Sun, Li, and Li]{ou2024absorbing}
Jingyang Ou, Shen Nie, Kaiwen Xue, Fengqi Zhu, Jiacheng Sun, Zhenguo Li, and
  Chongxuan Li.
\newblock Your absorbing discrete diffusion secretly models the conditional
  distributions of clean data.
\newblock In \emph{International Conference on Learning Representations}, 2025.

\bibitem[Patterson et~al.(2021)Patterson, Gonzalez, Le, Liang, Munguia,
  Rothchild, So, Texier, and Dean]{patterson2021carbon}
David Patterson, Joseph Gonzalez, Quoc Le, Chen Liang, Lluis-Miquel Munguia,
  Daniel Rothchild, David So, Maud Texier, and Jeff Dean.
\newblock Carbon emissions and large neural network training.
\newblock \emph{arXiv preprint arXiv:2104.10350}, 2021.

\bibitem[Peng et~al.(2023{\natexlab{a}})Peng, Li, He, Galley, and
  Gao]{peng2023instructiongpt4}
Baolin Peng, Chunyuan Li, Pengcheng He, Michel Galley, and Jianfeng Gao.
\newblock Instruction tuning with {GPT-4}.
\newblock \emph{arXiv preprint arXiv:2304.03277}, 2023{\natexlab{a}}.

\bibitem[Peng et~al.(2023{\natexlab{b}})Peng, Wu, Wei, Zhao, Yang, Liu, Xiong,
  Yang, Ni, Hu, Li, Zhang, Li, Ning, Wang, Zhang, Liu, Chau, Hu, and
  Cheng]{peng2023fp8lm}
Houwen Peng, Kan Wu, Yixuan Wei, Guoshuai Zhao, Yuxiang Yang, Ze~Liu, Yifan
  Xiong, Ziyue Yang, Bolin Ni, Jingcheng Hu, Ruihang Li, Miaosen Zhang, Chen
  Li, Jia Ning, Ruizhe Wang, Zheng Zhang, Shuguang Liu, Joe Chau, Han Hu, and
  Peng Cheng.
\newblock {FP8-LM}: Training {FP8} large language models.
\newblock \emph{arXiv preprint arXiv:2310.18313}, 2023{\natexlab{b}}.

\bibitem[Rein et~al.(2023)Rein, Hou, Stickland, Petty, Pang, Dirani, Michael,
  and Bowman]{rein2023gpqa}
David Rein, Betty~Li Hou, Asa~Cooper Stickland, Jackson Petty, Richard~Yuanzhe
  Pang, Julien Dirani, Julian Michael, and Samuel~R. Bowman.
\newblock {GPQA}: A graduate-level google-proof q\&a benchmark.
\newblock \emph{arXiv preprint arXiv:2311.12022}, 2023.

\bibitem[Sahoo et~al.(2024)Sahoo, Arriola, Schiff, Gokaslan, Marroquin, Chiu,
  Rush, and Kuleshov]{sahoo2024simple}
Subham~Sekhar Sahoo, Marianne Arriola, Yair Schiff, Aaron Gokaslan, Edgar
  Marroquin, Justin~T. Chiu, Alexander Rush, and Volodymyr Kuleshov.
\newblock Simple and effective masked diffusion language models.
\newblock \emph{arXiv preprint arXiv:2406.07524}, 2024.

\bibitem[Sahoo et~al.(2026)Sahoo, Chen, Pham, Geuter, Dwivedi, Pimpalkhute,
  Akhauri, Moreno, Yurochkin, Wang, Elhoushi, Dey, Bergsma, Hestness,
  Thickstun, Xing, and Liu]{sahoo2026uno}
Subham~Sekhar Sahoo, Lingjie Chen, Khiem Pham, Jonathan Geuter, Chaitanya
  Dwivedi, Varad Pimpalkhute, Yash Akhauri, Alexander Moreno, Mikhail
  Yurochkin, Zhenting Wang, Mostafa Elhoushi, Nolan Dey, Shane Bergsma, Joel
  Hestness, John Thickstun, Eric Xing, and Zhengzhong Liu.
\newblock Unlocking lossless speedups in {LLM}s via discrete diffusion.
\newblock \emph{arXiv preprint arXiv:2609.04010}, 2026.

\bibitem[Shi et~al.(2024)Shi, Han, Wang, Doucet, and
  Titsias]{shi2024simplified}
Jiaxin Shi, Kehang Han, Zhe Wang, Arnaud Doucet, and Michalis~K. Titsias.
\newblock Simplified and generalized masked diffusion for discrete data.
\newblock \emph{arXiv preprint arXiv:2406.04329}, 2024.

\bibitem[Strubell et~al.(2019)Strubell, Ganesh, and
  McCallum]{strubell2019energy}
Emma Strubell, Ananya Ganesh, and Andrew McCallum.
\newblock Energy and policy considerations for deep learning in {NLP}.
\newblock In \emph{Proceedings of the 57th Annual Meeting of the Association
  for Computational Linguistics}, pp.\  3645--3650, 2019.
\newblock \doi{10.18653/v1/P19-1355}.

\bibitem[Taori et~al.(2023)Taori, Gulrajani, Zhang, Dubois, Li, Guestrin,
  Liang, and Hashimoto]{taori2023alpaca}
Rohan Taori, Ishaan Gulrajani, Tianyi Zhang, Yann Dubois, Xuechen Li, Carlos
  Guestrin, Percy Liang, and Tatsunori~B. Hashimoto.
\newblock Stanford alpaca: An instruction-following {LLaMA} model.
\newblock Project release, 2023.
\newblock URL \url{https://crfm.stanford.edu/2023/03/13/alpaca.html}.

\bibitem[Turok et~al.(2026)Turok, De~Sa, and Kuleshov]{turok2026duel}
Gilad Turok, Chris De~Sa, and Volodymyr Kuleshov.
\newblock {DUEL}: Exact likelihood for masked diffusion via deterministic
  unmasking.
\newblock \emph{arXiv preprint arXiv:2603.01367}, 2026.

\bibitem[Wang et~al.(2026)Wang, Zhuang, Huang, and Zhang]{wang2026nara}
Shuaidi Wang, Zhan Zhuang, Ruping Huang, and Yu~Zhang.
\newblock {NaRA}: Noise-aware {LoRA} for parameter-efficient fine-tuning of
  diffusion {LLM}s.
\newblock \emph{arXiv preprint arXiv:2605.29716}, 2026.

\bibitem[Wang et~al.(2024)]{wang2024mmlupro}
Yubo Wang et~al.
\newblock {MMLU-Pro}: A more robust and challenging multi-task language
  understanding benchmark.
\newblock \emph{arXiv preprint arXiv:2406.01574}, 2024.

\bibitem[Wu et~al.(2025)Wu, Zhang, Xue, Diao, Fu, Liu, Molchanov, Luo, Han, and
  Xie]{wu2025fastdllmv2}
Chengyue Wu, Hao Zhang, Shuchen Xue, Shizhe Diao, Yonggan Fu, Zhijian Liu,
  Pavlo Molchanov, Ping Luo, Song Han, and Enze Xie.
\newblock Fast-{dLLM} v2: Efficient block-diffusion {LLM}.
\newblock \emph{arXiv preprint arXiv:2509.26328}, 2025.

\bibitem[Ye et~al.(2025)Ye, Xie, Zheng, Gao, Wu, Jiang, Li, and
  Kong]{ye2025dream}
Jiacheng Ye, Zhihui Xie, Lin Zheng, Jiahui Gao, Zirui Wu, Xin Jiang, Zhenguo
  Li, and Lingpeng Kong.
\newblock Dream 7b: Diffusion large language models.
\newblock \emph{arXiv preprint arXiv:2508.15487}, 2025.

\bibitem[Zellers et~al.(2019)Zellers, Holtzman, Bisk, Farhadi, and
  Choi]{zellers2019hellaswag}
Rowan Zellers, Ari Holtzman, Yonatan Bisk, Ali Farhadi, and Yejin Choi.
\newblock Hellaswag: Can a machine really finish your sentence?
\newblock In \emph{Proceedings of ACL}, 2019.

\bibitem[Zhou et~al.(2026)Zhou, Chen, Tong, and Song]{zhou2026dllm}
Zhanhui Zhou, Lingjie Chen, Hanghang Tong, and Dawn Song.
\newblock {dLLM}: Simple diffusion language modeling.
\newblock \emph{arXiv preprint arXiv:2602.22661}, 2026.

\end{thebibliography}
